\documentclass[journal]{IEEEtran}

\usepackage{graphicx}
\usepackage{amssymb}
\usepackage{amsmath}
\usepackage{float}
\usepackage{booktabs}
\usepackage{hyperref}

\begin{document}

\title{A Gesture Recognition System for Detecting Behavioral Patterns of ADHD}

\author{Miguel \'{A}ngel Bautista, Antonio Hern\'{a}ndez-Vela, Sergio Escalera, Laura Igual, Oriol Pujol\thanks{ Miguel \'{A}ngel Bautista, Antonio Hern\'{a}ndez-Vela, Sergio Escalera, Laura Igual and Oriol Pujol are with the Applied Mathematics and Analisis Department of the Universitat de Barcelona at Gran Via 585, 08007 Barcelona, Spain. They are also with the Computer Vision Center, Campus UAB, Edifici O 08193 Bellaterra, Barcelona, Spain, (email: \{mbautista,ahernandez,sescalera,ligual,opujol\}@ub.edu)}, Josep Moya\thanks{ Josep Moya is with the Parc Taul\'{i} Fundation at Parc Taulí, 1, 08208 Sabadell, Barcelona, Spain, (email:jmoya@tauli.cat)}, Ver\'{o}nica Violant, \thanks{Ver\'{o}nica Violant is with the Didactics and Educational Organization Department at University of Barcelona, P. de la Vall d?Hebron 171, 2ª planta, 08035 Barcelona, Spain, (email:vviolant@ub.edu)} and Mar\'ia Teresa Anguera \thanks{Mar\'{i}a Teresa Anguera is with the Department of Behavioral Sciences Methodologies at University of Barcelona. Psychology School, Campus de Mundet - Edifici Ponent Passeig de la Vall d'Hebron, 171
08035 Barcelona, Spain. (email:tanguera@ub.edu) }}

\maketitle

%\institute{$^1$ \at
%Dept. Matem\`{a}tica Aplicada i An\`{a}lisi\\ Universitat de Barcelona\\ Gran Via 585, 08007 Barcelona, Spain.\\
% \email{\{mbautista,ahernandez,sescalera,ligual,opujol\}@ub.edu}\\
%$^2$ \at
%Centre de Visi\'{o} per Computador\\ Campus UAB, Edifici O\\ 08193 Bellaterra, Barcelona, Spain.\\
%$^3$ \at  Servicio de Salud Mental del Parc Taulí, Sabadell\\ \email{jmoya@tauli.cat}\\
%$^4$ \at Facultad de Pedagog\'{i}a, Universitat de Barcelona, \email{vviolant@ub.edu}}

\begin{abstract}

We present an application of gesture recognition using an extension of Dynamic Time Warping (DTW) to recognize behavioural patterns of Attention Deficit Hyperactivity Disorder (ADHD). We propose an extension of DTW using one-class classifiers in order to be able to encode the variability of a gesture category, and thus, perform an alignment between a gesture sample and a gesture class. We model the set of gesture samples of a certain gesture category using either GMMs or an approximation of Convex Hulls. Thus, we add a theoretical contribution to classical warping path in DTW by including local modeling of intra-class gesture variability. This methodology is applied in a clinical context, detecting a group of ADHD behavioural patterns defined by experts in psychology/psychiatry, to provide support to clinicians in the diagnose procedure. The proposed methodology is tested on a novel multi-modal dataset (RGB plus Depth) of ADHD children recordings with behavioural patterns. We obtain satisfying results when compared to standard state-of-the-art approaches in the DTW context.
\end{abstract}


\section{Introduction}

%COMPUTER VISION SYSTEMS TO RECOGNIZE BEHAVIOUR AND ADHD

Nowadays, human gesture recognition is one of the most challenging
tasks in computer vision. Due to the large number of potential
applications involving human gesture recognition in fields like
surveillance~\cite{Hampapur}, sign language
recognition~\cite{sign}, or clinical assistance~\cite{pentland}
among others, there is a large and active research community
devoted to deal with this problem. Current methodologies have shown
preliminary results on very simple scenarios, but they are still
far from human performance.

In the gesture recognition field there exists a wide number of
methods based on dynamic programming algorithms for both alignment
and clustering of temporal series~\cite{haca}. Probabilistic
methods such as Hidden Markov Models (HMM) or Conditional Random
Fields (CRF) are also very usual in the literature~\cite{sign}. Nevertheless, one
of the most common methods for Human Gesture Recognition is
Dynamic Time Warping (DTW)~\cite{Reyes,bautistadtw}. It offers a simple
yet effective temporal alignment between sequences of different
lengths. However, the application of such methods to gesture
detection in complex scenarios becomes a hard task due to the high
variability of the environmental conditions among different domains. Some
common problems are: wide range of human pose configurations,
influence of background, continuity of human movements,
spontaneity of human actions, speed, appearance of unexpected
objects, illumination changes, partial occlusions, or different
points of view, just to mention a few. These effects can cause
dramatic changes in the description of a certain gesture,
generating a great intra-class variability. In this sense, since
usual DTW is applied between a sequence and a single pattern, it
fails to take into account such variability.

In addition, the release of the Microsoft Kinect$^{TM}$ sensor in late 2010 has
allowed an easy and inexpensive access to synchronized depth
imaging with standard video data. This data combines both sources
into what is commonly named RGB-D images (RGB plus Depth).
This data fusion, very welcomed by the computer vision community, has
reduced the burden of the first steps in many pipelines devoted to
image or object segmentation and opened new questions such as how
this data can be effectively described and fused. This depth information has been particularly
exploited for human body segmentation and tracking.
Shotton~\cite{Shotton} introduced one of the greatest advances in
the extraction of the human body pose using RGB-D, which is
provided as part of the Kinect$^{TM}$ human recognition framework.
The method is based on inferring pixel label probabilities through Random Forest from
learned offsets of depth features. Girshick and Shotton~\cite{ShottonICCV}
proposed later a different approach in which they directly regress
the positions of the body joints, without the need of an intermediate
pixel-wise body limb classification as in~\cite{Shotton}.
The extraction of body pose information opens the door to develop more accurate gesture
recognition methodologies.

In particular, there is a growing interest in the application 
of gesture recognition methods in the clinical context. Concretely, gesture recognition methods can be 
even more valuable on psychological or psychiatric scenarios where the diagnostic of a certain disease is based on 
the interpretation of certain behavioural patterns of the subject. Up to date, video sequences were analysed on a frame-by-frame fashion by experts which were typically trained for several months to achieve a good performance on the analysis. Of course, this situation is not applicable to large amounts of data since it is a very time consuming procedure and its automatization is highly desirable. Specifically, the case of Attention Deficit Hyperactivity Disorder (ADHD) is one of the most notable scenarios, since it is the most commonly studied and diagnosed psychiatric disorder in childhood, globally affecting about 5 percent of children \cite{tdah1}. In this line of research some works can be found in literature \cite{introtdah1,introtdah2}, which develop tools to assist children with autism-related disorders. Nevertheless, one of the main problems that clinicians experiment when diagnosing ADHD is the huge subjective component of the interpretation of symptoms, because their definition is either ambiguous or inaccurate. In this sense, an objective gesture recognition tool which is able to detect behavioural patterns defined by a set of psychiatric/psychological experts will be of great value in order to help the clinicians with ADHD diagnose. This work pretends to be a study on a concrete set of ADHD patterns, which aims to be extended in future works.

%proposal
We propose to use an extension of the DTW method, that is
able to perform an alignment between a sequence and a set of $N$
pattern samples from the same gesture category. The variance
caused by environmental factors is modelled using either a Gaussian
Mixture Model (GMM)~\cite{gmm} or an approximation of a Convex Hull \cite{piero}. Consequently, the distance metric
used in the DTW framework is redefined in order to provide a probability-based measure. The proposed method is evaluated in a novel ADHD behavioural pattern dataset, in which both subject diagnosed with ADHD and a control group where recorded in a class-room environment, obtaining satisfying results. Our list of contributions is as follows: i) An extension of classical DTW by modelling the intra-class variability of gestures is proposed. ii) GMMs and approximated Convex Hulls are embedded in the DTW by defining novel distances.  iii) A novel multi-modal ADHD behavioural pattern dataset is presented. iv) We test our proposal in the novel ADHD behavioural patterns dataset obtaining very satisfying results.

The rest of the paper is structured as follows: Section \ref{method} presents the Gesture Recognition proposal. Section \ref{experiments}  presents a novel ADHD dataset and shows the experimental results on a novel ADHD behavioural pattern dataset. Finally, Section \ref{conclusions} summarizes the conclusions.

\section{Definition of ADHD Behavioural Patterns and Feature Extraction}\label{method}

We split the methodology of the proposal in different stages. First, we define the ADHD behavioural patterns to be learnt. Second, the considered set of multi-modal features for each frame is described, and finally, the novel DTW extension based on GMM and Convex Hull modelling is presented.

\subsection{Definition of ADHD Behavioural Patterns}\label{indicators}

Attention Deficit Hyperactivity Disorder (ADHD) is one of the most common childhood disorders and can continue through adolescence and adulthood. Symptoms include difficulty staying focused and paying attention, difficulty controlling behaviour, and hyperactivity. ADHD has three subtypes, defined by DSM IV and CIE X \cite{dsm,cie}:
\begin{enumerate}

\item Predominantly hyperactive-impulsive

\item Predominantly inattentive

\item Combined hyperactive-impulsive and inattentive

\end{enumerate}

In addition children who have symptoms of inattention may:

\begin{itemize}
\item Be easily distracted, miss details, forget things, and frequently switch from one activity to another.
\item Have difficulty focusing on one task.
\item Become bored with a task after only a few minutes, unless they are doing something enjoyable.
\item Have difficulty focusing attention on organizing and completing a task or learning something new.

\end{itemize}

Children who have symptoms of hyperactivity may:

\begin{itemize}

\item Fidget and squirm in their seats.
\item Dash around, touching or playing with anything and everything in sight.
\item Have trouble sitting still during dinner, school, and story time.
\item Be constantly in motion.

\end{itemize}

Children who have symptoms of impulsiveness may:

\begin{itemize}

\item Have difficulty waiting for things they want or waiting their turns in games.
\item Often interrupt conversations or other activities.
\end{itemize}

%Parrafo de elegir los indicadores
In order to develop a system that automatically detects ADHD behavioural patterns, first we have to define a set of ADHD behavioural patterns (gestures to detect) that are both objective and descriptive yet discriminable. In other words, the set of patterns has to be descriptive enough to provide an ADHD profile of the subject, and simple enough in order to be able to automatize the detection.

\begin{figure}[!htbp]
\centering
\begin{tabular}{cc}
\includegraphics[width=3.75 cm]{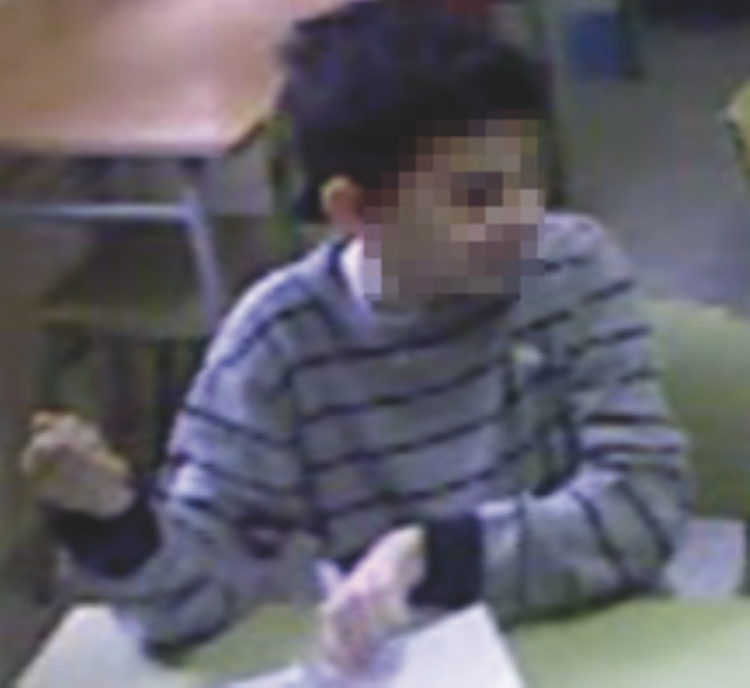} & \includegraphics[width=3.75 cm]{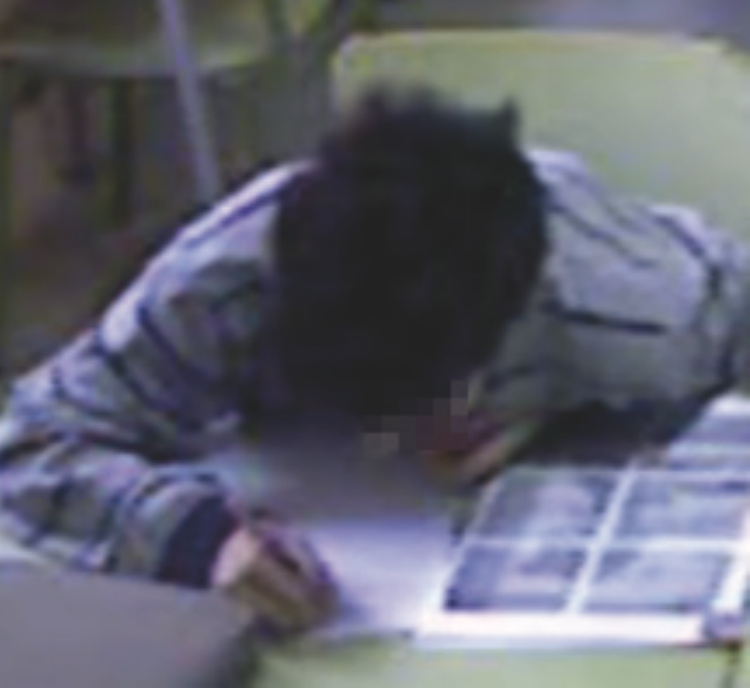} \\
(a) & (b)\\
\includegraphics[width=3.75 cm]{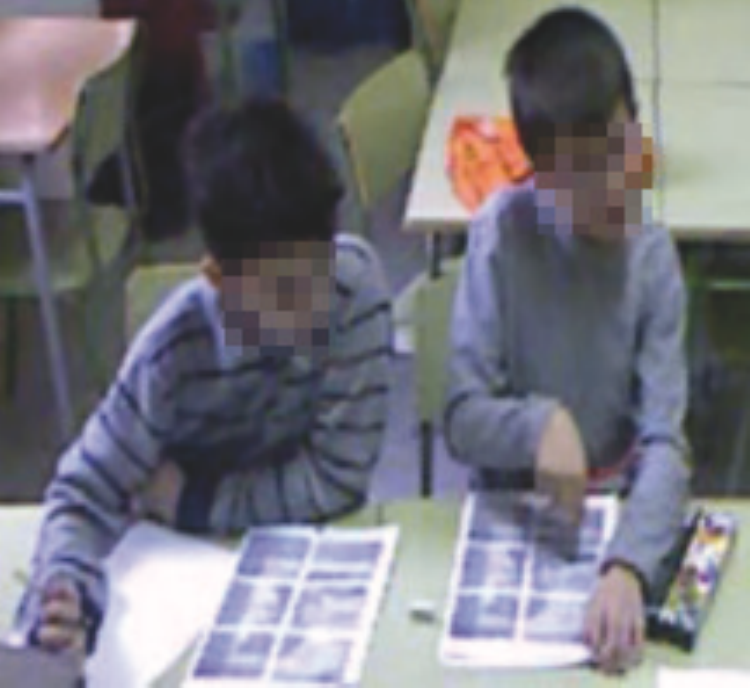} & \includegraphics[width=3.75 cm]{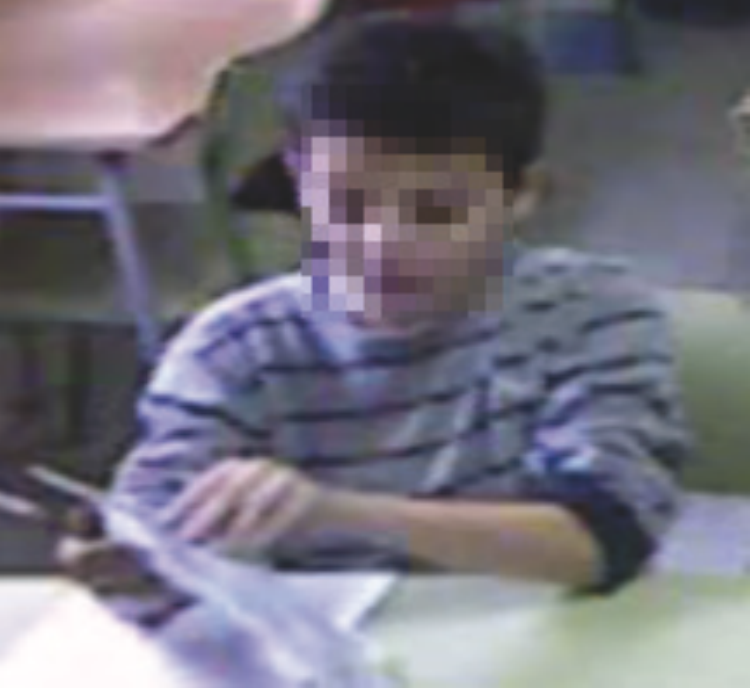} \\
(c) & (d) \\
\end{tabular}
\caption{(a) An example of a the head turning behavioural pattern. (b) Torso in table pattern example, notice how the torso of the subject is completely laid on the table . (c) Sample of a class mate invasion in which the left subject invades the right subject space. (d) Movement behavioural pattern sample.}
\label{fig::indicators}
\end{figure}

In order to define the behavioural patterns to be automatically detected, an analysis of the context in which the video sequences take place has to be performed. Taking into account that video sequences were recorded in a school class context, including mathematical exercises and computer gaming, with no disturbing events taking place, the set of defined ADHD behavioural patterns is the following (an example is shown in Figure \ref{fig::indicators}):

\begin{itemize}

\item \textbf{Head turning behavioural pattern}

The definition of this behavioural pattern takes its reason from the different symptoms in the \textit{inattention} branch. Behaviours like \textit{be easily distracted, miss details, forget things, and frequently switch from one activity to another.} or  \textit{have difficulty focusing on one thing} have a close relationship with turning the head from the goal task to other unrelated task. Therefore, this indicator is defined as a head turn to either right or left sides.

\item \textbf{Torso in table behavioural pattern}

The Torso in table behavioural pattern is related to \textit{hyperactive} symptoms such \textit{fidget and squirm in their seats} and \textit{have trouble sitting still during dinner, school, and story time}.

\item \textbf{Classmate's desk invasion behavioural pattern}

This behavioural pattern takes is root from the \textit{impulsive} symptoms like \textit{often interrupt conversations or others' activities} or \textit{have difficulty waiting for things they want or waiting their turns in games}.

\item \textbf{Movement with/without a pattern behavioural pattern}

The last pattern aims to provide a detection for those symptoms across all ADHD branches (inattentiveness, hyperactivity and impulsiveness) that involve a high quantity of motion.

\end{itemize}

This set of behavioural patterns is representative enough of the different symptoms of ADHD and provides a generalization analysis of the feasibility of our approach for supporting diagnosis.

\subsection{Image Acquisition, Pre-processing and Feature Extraction}

%Se debe extender
We use the Kinect$^{\copyright}$ sensor in order to capture video sequences in which subjects diagnosed with ADHD and subjects not diagnosed with ADHD (control group) were recorded. In this sense, we use the depth information provided by the Kinect$^{\copyright}$ sensor to obtain a segmentation of the subjects in the scene obtaining a complete segmentation of their upper-body limbs.

Given a frame $I^t, t=1,..,T$, the corresponding segmentation $S^t$ on the depth map is computed by the OTSU method \cite{otsu}, keeping the biggest convex unconnected components in relation with the number of subjects appearing in the scene. In other words, if three subjects appear on the scene, the three biggest components were kept. Otherwise, if two subjects appear on the scene, the biggest components are kept as the segmentation. Moreover, Random Forest segmentation is applied over the foreground objects \cite{Shotton} in order to segment the regions corresponding to different subjects.

\subsubsection{Head Turning Behavioural Pattern Feature}

\begin{figure}[htbp]
\begin{center}
\includegraphics[width=8.85 cm]{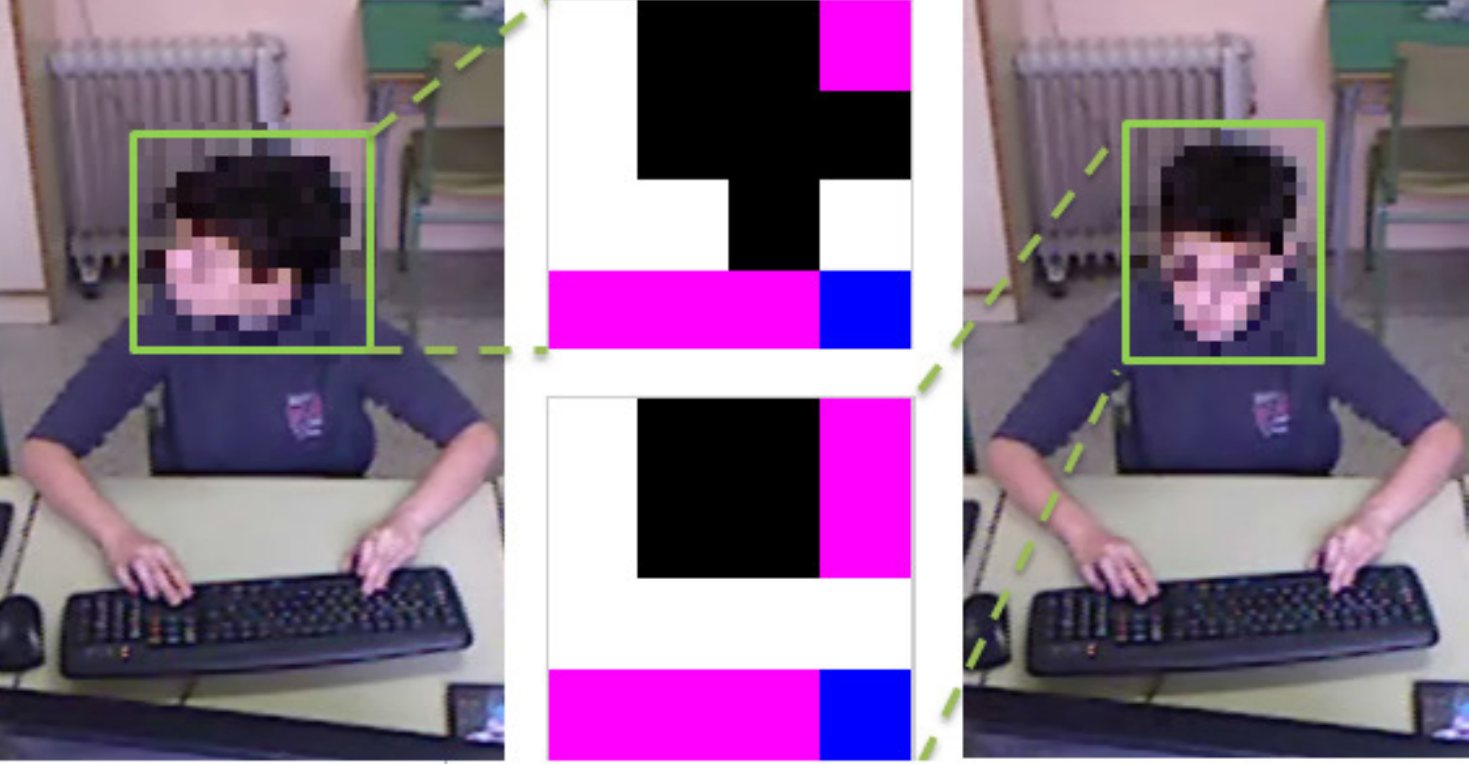}
\end{center}
\caption{ Color descriptor for the 'Head turning' behavioural pattern. Image in first column shows a subject turning the head, while in the image at the last column shows a frontal face. Bounding boxes are overlaid in green color. Images in the central column show the respective color naming descriptors. They are composed by $4 \times 4$ cells, each one of them containing a color name label.} \label{fig::featcabeza}
\end{figure}

The features for the head rotation detection are computed for each frame $t$ as follows. First of all, we obtain the bounding box $B^t$ containing the head, by means of GrabCut segmentation \cite{GrabCut}. As GrabCut is a semi-automatic method, a manual bounding box has to be provided by the user at the first frame. With the resulting segmentation mask, the bounding box for that frame can be easily computed. Additionally, some morphological operations are applied on the segmentation mask in order to initialize the segmentation of the following frame, as in \cite{toni1}.
Once a bounding box $B^t$ is detected for one frame, a color-based descriptor $FHead^t$ is extracted from the pixels inside it. The bounding box is firstly divided in $\bar{O}\times O$ cells, and each one of them is described with a label $\gamma \in \{1,...,G\}$ corresponding to the most frequent color as follows:

\begin{eqnarray}
FHead^{t}_{i,j} =\arg\max_{l \in \gamma} \left(\sum_{\mathbf{x} \in B^t_{i,j}} \delta(\mathrm{ColorName}(\mathbf{x}) - l)\right), \nonumber \\ \forall i \in 1,..,\bar{O},  \nonumber \\ \forall j \in 1,..,O,
\end{eqnarray}

where $B^t_{i,j}$ is the $(i,j)$-th cell of the head bounding box at time $t$. In addition, $\mathrm{ColorName}(\mathbf{x})$ is a function which returns the color name of an RGB pixel $\mathbf{x}$,
and $\delta(\cdot)$ is a Dirac delta function.
The Color-naming data with $G=11$ basic colors (red, orange, brown, yellow, green,
blue, purple, pink, white, grey, black) presented in \cite{toni2} has been used. An example of the feature computation procedure is show in Figure \ref{fig::featcabeza}.

\subsubsection{Torso on Desk Behavioural Pattern Feature}

The torso on desk feature computes the relative distance of the subject's torso to the desk,
in order to provide a measure of how close the torso is in relation to the desk. In this sense,
this distance is computed as the Euclidean distance of the top pixel of the head to the closest
desk pixel. This distance can be easily computed by finding the uppermost pixel
$x^{top} = \{x_i | (x_i, y_i)\in S^t, (x_j, y_j)\in S^t, y_i\leq y_j, \forall i \neq j \}$ in the
segmentation mask $S^t$ of the subject, and its corresponding lowermost pixel in vertical direction
$x^{bot} = \{x_i | (x_i, y_i)\in S^t, (x_j, y_j)\in S^t, y_i\geq y_j, \forall i \neq j \}$:
\begin{equation}
FTorso^t = \|\mathbf{x}^{top} - \mathbf{x}^{bot}\|_2.
\end{equation}

An example of the feature calculation is shown in Figure \ref{fig::feattorso}.

\subsubsection{Classmate's Desk invasion feature}

In order to compute the Classmate's Desk Invasion feature, we also use the segmentation mask $S^t$.
For a given subject, the feature is basically defined as the minimum distance
between the pixels in the subject's unconnected components of the mask $S^t$, and
the pixels in the neighbour classmate's components $S^t_{ne}$ ($ne=1,2$ in our case):
\begin{equation}
FInv^t = \min_{ne \in N} \left( \min_{\mathbf{x}_n \in S^t_n} \left( \min_{\mathbf{x}_s \in S^t} \| \mathbf{x}_s - \mathbf{x}_n  \|_2 \right) \right).
\end{equation}

\begin{figure}[htbp]
\begin{center}
\includegraphics[width=8.85 cm]{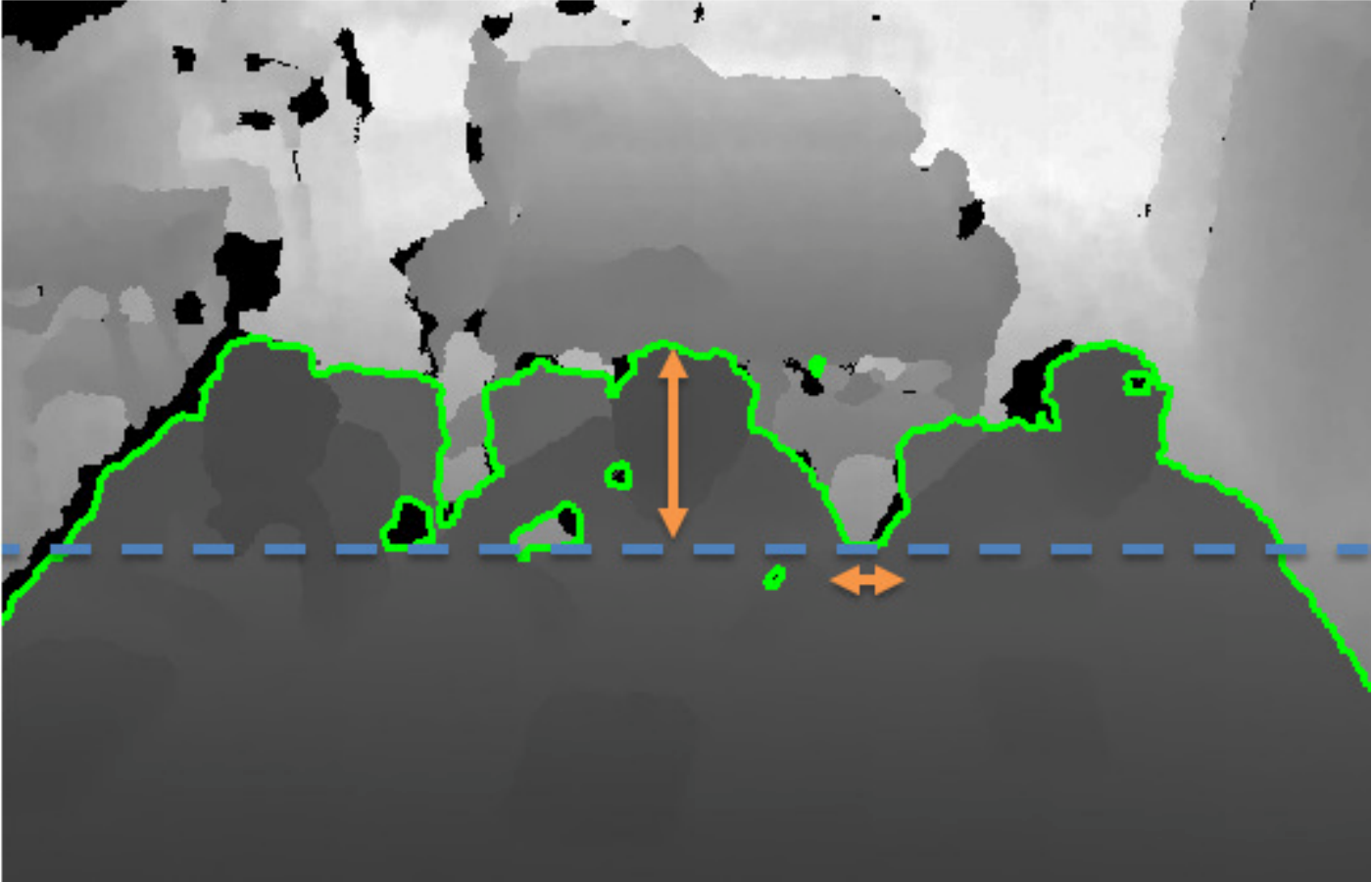}
\end{center}
\caption{  Distances computed using the segmentation of the depth image. Green contour indicates the boundary of the segmentation mask. Blue dashed line shows the table limit. Vertical and horizontal orange arrows show the distances computed for the "Torso in table" and 'Classmate´s desk invasion' behavioural patterns, respectively.} \label{fig::feattorso}
\end{figure}

An example of this computation is shown in Figure \ref{fig::feattorso}.
%To compute this feature we use the segmentation provided by the Kinect$^{\copyright}$ sensor, in order to obtain a segmentation of each subject appearing in the scene. Then, we compute the distance between unconnected convex components obtained from the depth segmentation.

\subsubsection{Movement with/without a pattern feature}
In the movement with/without a pattern feature we want to describe general movements of the subject, so we first compute the
optical flow \cite{takeo} between current and next frames. Then, we compute the average optical flow magnitude over 
the pixels belonging to the segmentation mask of the subject:
\begin{equation}
FMov^t = \frac{1}{|S^t|} \sum_{\mathbf{x} \in S^t} \sqrt{u_{\mathbf{x}}^2 + v_{\mathbf{x}}^2},
\end{equation}
where $u_{\mathbf{x}}$ and $v_{\mathbf{x}}$ are the components of the flow vector between current frame
$I^t$ and $I^{t+1}$, and $|\cdot|$ computes the number of elements of the set.

\section{Dynamic Time Warping based on one-class classifiers}

The original DTW algorithm was defined to match temporal
distortions between two models, finding an alignment/warping path
between the two time series $Q=\{q_1,..,q_n\}$ and
$C=\{c_1,..,c_m\}$. In order to align these two sequences, a
$M_{m\times n}$ matrix is designed, where the position $(i,j)$ of
the matrix contains the alignment cost between $c_i$ and $q_j$.
Then, a warping path of length $\tau$ is defined as a set of
contiguous matrix elements, defining a mapping between $C$ and
$Q$: $W=\{w_1,..,w_\tau \}$, where $w_i$ indexes a position in the
cost matrix. This warping path is typically subjected to several
constraints:

\begin{normalsize}

\emph{Boundary conditions:} $w_1=(1,1)$ and $w_\tau=(m,n)$.

\emph{Continuity and monotonicity:} Given $w_{\tau'-1}=(a',b')$, then
$w_{\tau'}=(a,b)$, $a-a'\leq 1$ and $b-b'\leq 1$. This condition
forces the points in $W$ to be monotonically spaced in time.
\end{normalsize}

We are generally interested in the final warping path that,
satisfying these conditions, minimizes the warping cost:

\begin{equation}
DTW(M)= \min_{W} \left\{\frac{M(w_\tau)}{\tau} \right\},
\end{equation}

where $\tau$ compensates the different lengths of the warping paths.
This path can be found very efficiently using dynamic programming.
The cost at a certain position $M(i,j)$ can be found as the
composition of the Euclidean distance $d(i,j)$ between the feature
vectors of the sequences $c_i$ and $q_j$ and the minimum cost of
the adjacent elements of the cost matrix up to that point, i.e.:
$M(i,j)=d(i,j)+\min\{M(i-1,j-1),M(i-1,j),M(i,j-1)\}$.

Given the streaming nature of our problem, the input vector $Q$
has no definite length and may contain several occurrences a gesture class, namely $C$. At that point the system considers that there is correspondence between the current block $k$ in $Q$
and a gesture if satisfying the following condition,
$M(m,k)<\beta, k\in [1,..,\infty]$
for a given cost threshold $\beta$.

This threshold is estimated in advance using leave-one-out
cross-validation strategy on the training set. This involves using a single
observation from the original sample as the validation data, and
the remaining observations as the training data. This is repeated
such that each observation in the sample is used once as the
validation data. At each iteration, we evaluate the similarity
value between the candidate and the rest of the training set.
Finally, we choose the threshold value which is associated with the
largest number of hits.

Once the threshold is defined and a possible end of pattern of gesture is detected, the
working path $W$ can be found through backtracking of the minimum
path from $M(m,k)$ to $M(0,z)$, being $z$ the instant of time in
$Q$ where the gesture begins. Note that
$d(i,j)$ is the cost function which measures the difference among
our descriptors $c_i$ and $q_j$.

An example of a begin-end gesture recognition
together with the warping path estimation is shown in
Figure~\ref{fig:analyzer}.

\subsection{Handling temporal deformation in sequences}

Consider a training set of $N$ sequences $\{S_1,S_2,\ldots,S_N\}$, where all sequences belong to a certain gesture class.
Then, each sequence $S_g$ is composed by a  set of feature vectors at
each time $t$, $S_g=\{s^g_1,\ldots, s^g_{L_g}\}$, where $L_g$ is the length in frames of sequence
$S_g$. Let us assume that sequences are ordered according to their
length, so that $L_{g-1}\leq L_g\leq L_{g+1}, \forall g\in
[2,..,N-1]$, and the median length sequence is
$\bar{S}=S_{\lceil\frac{N}{2}\rceil}$.  This sequence is used as a
reference, and the rest of the sequences are aligned with respect to it using the
classical Dynamic Time Warping with Euclidean distance, in order to avoid the temporal
deformations of different samples from the same gesture category.
Therefore, after the alignment process, all sequences have length
$L_{\lceil\frac{N}{2}\rceil}$. We define the set of warped sequences as
$\{\tilde{S}_1,\tilde{S}_2,\ldots,\tilde{S}_N\}$.

\begin{figure}[htbp]
\begin{center}
\includegraphics[width=8.85 cm]{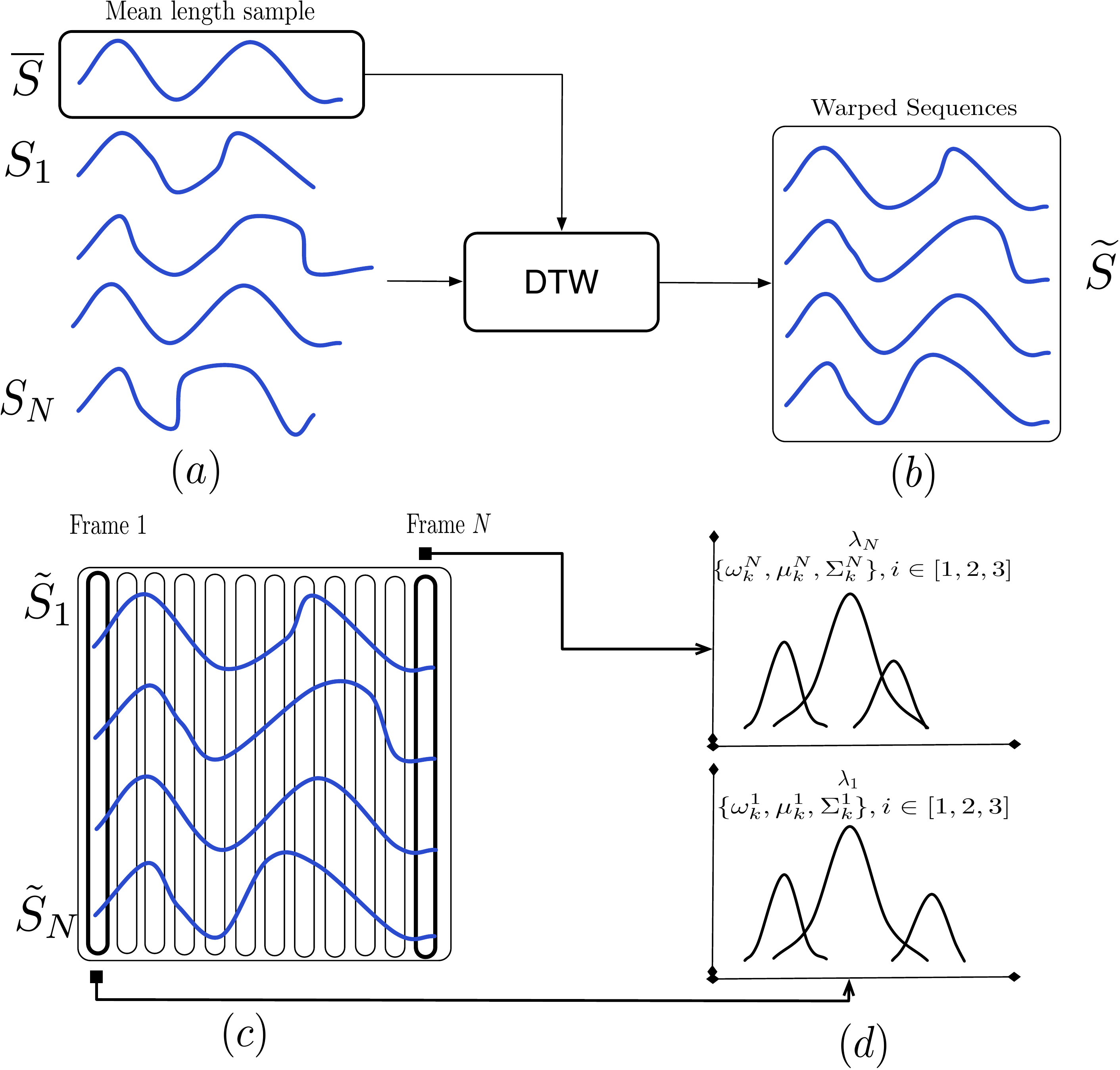}
\end{center}
\caption{(a) Different sample sequences of a certain gesture category and the mean length sample. (b) Alignment of all samples with the mean length sample by means of Euclidean DTW. (c) Warped sequences set $\tilde{S}$ from which each set of $t$-th elements among all sequences are modelled. (d) Gaussian Mixture Model learning with 3 components.}
\label{fig::preprocess}
\end{figure}

Once all samples are aligned, the feature vectors corresponding to a certain time $t$ among all sequences
$\tilde s^{g}_{t} \  \forall g \in [1,\dots,N]$ are modelled by means of one-class classifiers (i.e GMMs) in order to encode intra-class variability. An example of the process using GMMs is shown in Figure \ref{fig::preprocess}.

\subsection{Embedding One-Class Classifiers in DTW}

In the classical DTW, a pattern and a sequence are aligned using a
distance metric, such as the Euclidean distance. Since our
pattern is modelled by means of one-class models, if we want
to use the principles of DTW, the distance needs to be redefined. Next, we propose two cost distances, one based on GMM and the other on approximated Convex Hull.

\subsubsection{Gaussian Mixture Models} \label{GMM}

We propose to use Gaussian Mixture Models (GMM) to learn the features among all sequence samples (of a gesture category) at a certain time $t$, $\tilde s^{g}_{t} \  \forall g \in [1,\dots,N]$. Since after the alignment step all the sequences have the same length, $L_{\lceil\frac{N}{2}\rceil}$, we learn $L_{\lceil\frac{N}{2}\rceil}$ GMMs, one per each component.

In this sense, a $G-$component Gaussian
Mixture Model, is defined as, $\lambda_t=\{\alpha^{t}_{k},
\mu^{t}_{k},\Sigma^{t}_{k} \},\quad k = 1,..,G$, where $\alpha$ is the
mixing value and $\mu$ and $\Sigma$ are the parameters of each of
the $G$ Gaussian models in the mixture. As a result, each one of
the GMMs that model each set of t$-th$ components 
$\tilde{s}_t$, among all warped sequence samples, is defined as follows:
 
\begin{equation}
p(\tilde{s}_t)=\sum\limits_{k=1}^{G}  \alpha_k \cdot e^{-\frac{1}{2} (q-\mu_k)^{T}\cdot\Sigma_{k}^{-1}\cdot(q-\mu_k)}.
\label{eq:GMM}
\end{equation}

The resulting model is composed by a set of $L_{\lceil\frac{N}{2}\rceil}$ GMMs corresponding to the modelling of each one of the component elements of the warped sequence $\tilde{s}_t$ for each gesture pattern.

In this paper we consider a soft-distance based on the probability
of a point belonging to each one of the $G$ components in the GMM,
i.e., the posterior probability of $q \in Q$ is obtained according to Equation
\ref{eq:GMM}. In addition, since $\sum\limits_{1}^{k}
\alpha_k=1$, we can compute the probability of $x$ belonging to
the whole GMM $\lambda$ as the following:

\begin{equation}
P_{\text{GMM}}(q,\lambda)=\sum\limits_{k=1}^{M} \alpha_k \cdot P(q)_{k},
\end{equation}
\begin{equation}
P(x)_{k}= e^{-\frac{1}{2} (x-\mu_k)^{T}\cdot\Sigma_{k}^{-1}\cdot(x-\mu_k)},
\end{equation}

which is the sum of the weighted posterior probability of each component. However, an
additional step is required since the standard DTW algorithm is
conceived for distances instead of similarity measures. In this
sense, we use a soft-distance based measure of the probability,
which is defined as:

\begin{equation}\label{distance}
D(x,\lambda)=e^{-P_{\text{GMM}}(x,\lambda)}.
\end{equation}

\begin{figure}[htbp]
\begin{center}
\includegraphics[width=8.85 cm]{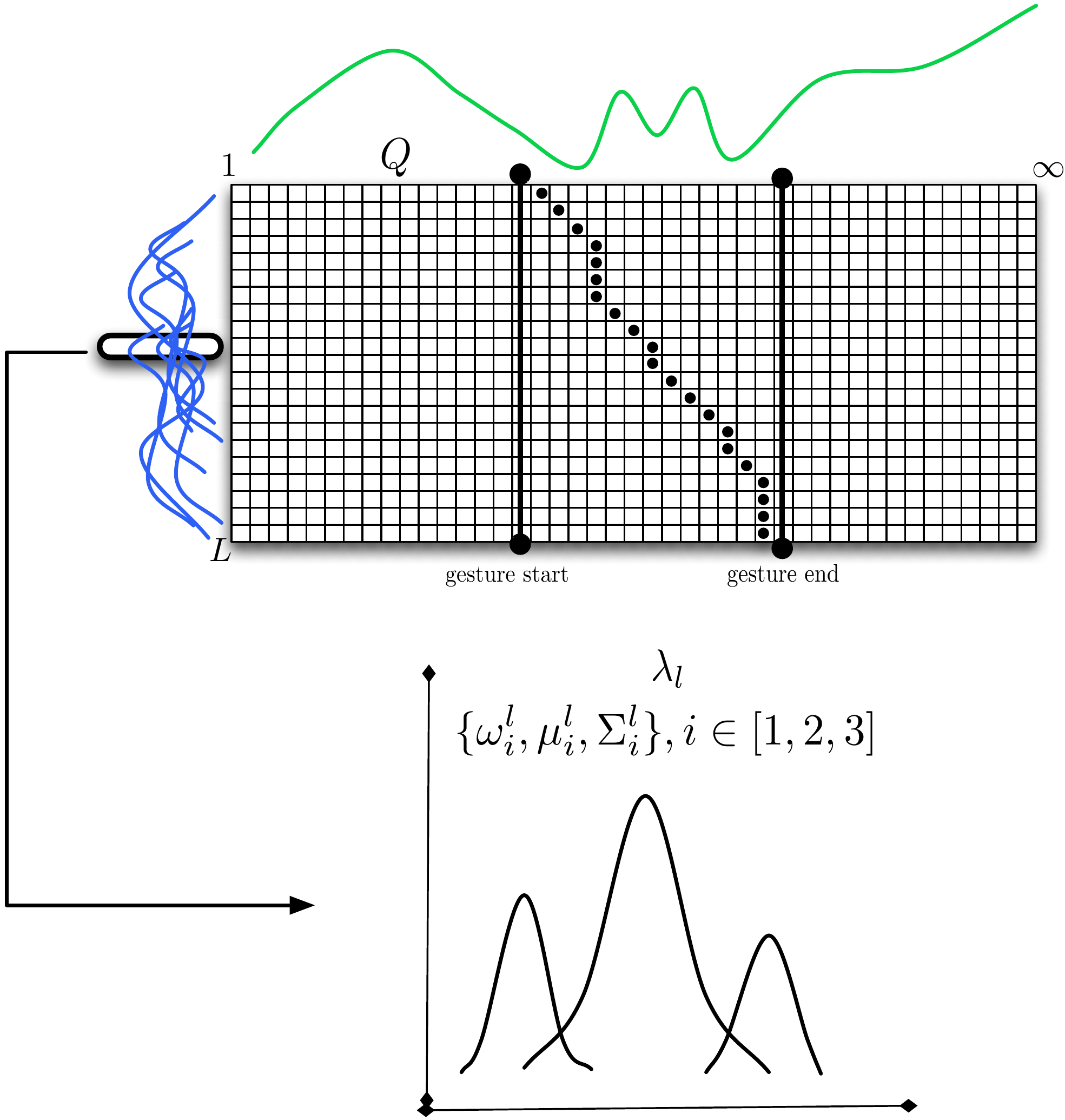}
\end{center}
\caption{Begin-end of gesture recognition of a gesture pattern in
an infinite sequence $Q$ using the probability-based DTW. Note
that different samples of the same gesture category are modelled
with a GMM and this model is used to provide a probability-based
distance. In this sense, each cell of $M$ will contain the
accumulative $D$ distance.} \label{fig:analyzer}
\end{figure}

An example of the use of GMMs framework to detect a given gesture is shown in Figure \ref{fig:analyzer}.

\subsubsection{Convex Hulls and Approximate Convex Polytope decision Ensemble} \label{CH}

In addition to the use of GMMs as One-class classifiers, we also propose to use Convex Hulls to model the set of features, $\tilde s^{g}_{t} \  \forall g \in [1,\dots,N]$. The underlying idea of Convex Hulls is to model the boundary of the set of points defining the problem.  If the boundary encloses a convex area, then the convex hull, defined as the minimal convex set containing all the training points, provides a good general tool for modelling the target class, which in our case will be the set of features of all sequence samples at a certain time.

The convex hull of a set $\mathcal{C} \subseteq  \mathbb{R}^\varrho$, denoted ${\bf conv} \,\mathcal{C}$, is defined as the smallest convex set that contains $\mathcal{C}$ 
and is defined as the set of all convex combinations of points in $\mathcal{C}$:
\begin{equation}
{\bf conv} \,\mathcal{C} = \{\theta_1 x_1 + \cdots + \theta_m x_m \, | \, x_i \in \mathcal{C}, \theta_i \geq 0, \forall i; \sum_i \theta_i = 1\}
\end{equation}
In this scenario, the one-class classification task is reduced to the problem of knowing if test data lie inside or outside the hull.
Although the convex hull provides a compact representation of the data, a small amount of outliers may lead to very different shapes of the convex polytope. Thus, a decision using these structures is prone to over-fitting. In \cite{piero}, the authors show that it is useful to define a parametrized set of convex polytopes associated with the original convex hull of the training data. 
This set of polytopes are shrunk/enlarged versions of the original convex hull governed by a parameter $\varphi$. 
The goal of this family of polytopes is to define the degree of robustness to outliers.  The parameter $\varphi$ defines a constant shrinking ($ -\|\wp-\varsigma\|\leq \varphi \leq 0$) or enlargement ($\alpha \geq 0$) of the convex structure with respect to the center $c$. If $\varphi=0$ then $\wp_{0}= {\bf conv}\, \mathcal{C}$.

However, the creation of high-dimensional convex hulls is computationally intensive. In general, the cost for computing  a $\varrho$-dimensional convex hull on $N$ data examples is $\mathcal{O}(N^{\lfloor{\varrho/2}\rfloor+1})$. 
This cost is prohibitive in time and memory and, for the classification task, only checking if a point lies inside the multidimensional structure is needed. Instead, we propose to use the Approximate convex Polytope decision Ensemble (APE) of \cite{piero}. This method consists in approximating the decision made using the extended convex polytope in the original $\varrho$-dimensional space by aggregating a set of $F$ decisions made on low-dimensional random projections of the data.

Since the projection matrix is created at random, the resulting space does not preserve the norm of the original space. Hence, a constant value of the parameter $\varphi$ in the original space corresponds to a set of values $\gamma_i$ in the projected one. As a result, the low-dimensional approximation of the expanded polytope is defined by the set of vertices as follows:

\begin{equation}
\bar{\wp}^{\varphi} : \{\bar{\wp_i}+ \omega_i \frac{(\bar{\wp_i}-\bar{\varsigma})}{\|\bar{\wp_i}-\bar{\varsigma}\|} \}, i=1,..,N,
\label{eq:extended}
\end{equation}

where  $\bar{\varsigma} = \rho \varsigma$ represents the projected center, $\bar{\wp_i}$  is the set of vertices belonging to the convex hull of the projected data and $\gamma_i$ is defined as follows:

\begin{equation}
 \omega_i =\frac{(\wp_i - \varsigma)^T \rho^T {\rho}  (\wp_i - \varsigma)}{{\| \wp_i - \varsigma \|}}  \alpha,
 \label{eq:gamma}
\end{equation}

where $\rho$ is the random projection matrix, $\varsigma$ is the center and $\wp_i$ is the $i$th vertex of the convex hull in the original space. Note that there exist a different expansion factor for each vertex ${\wp_i}$ belonging to the projected convex hull. Thus, we defined an APE model as:

\begin{equation}
\psi=\{\bar{\wp}^{\varphi}_{f}\},
\end{equation}

where $f \in [1,\dots,F]$, and $F$ is the number of total random projections used to approximate the original convex hull. In this sense, to obtain the probability of a point belonging to the extended/shrunken convex polytope ensemble we compute the proportion of low-dimensional random projections in which the testing point $q$ lies inside the extended convex polytope. In this sense, we get an approximate measure of how probable is the point to be inside the original Convex Hull. The calculation of the proportion is as follows:

\begin{equation}\label{eq::propCH}
P_{APE}(q,\psi)=\frac{\sum\limits_{f=1}^{F} q \in  \text{conv} \; \wp^{\varphi}_f}{F}.
\end{equation}

Following the same scheme used with GMM, we compute a soft distance based on the proportion of random projections in which the testing point $q$ lies inside the extended convex polytope. This soft-distance is defined as follows,
\begin{equation}
D(q,\psi)=e^{-P_{\psi}(q,\psi)}.
\end{equation}

\begin{figure}[htbp]
\begin{center}
\includegraphics[width=8.85 cm]{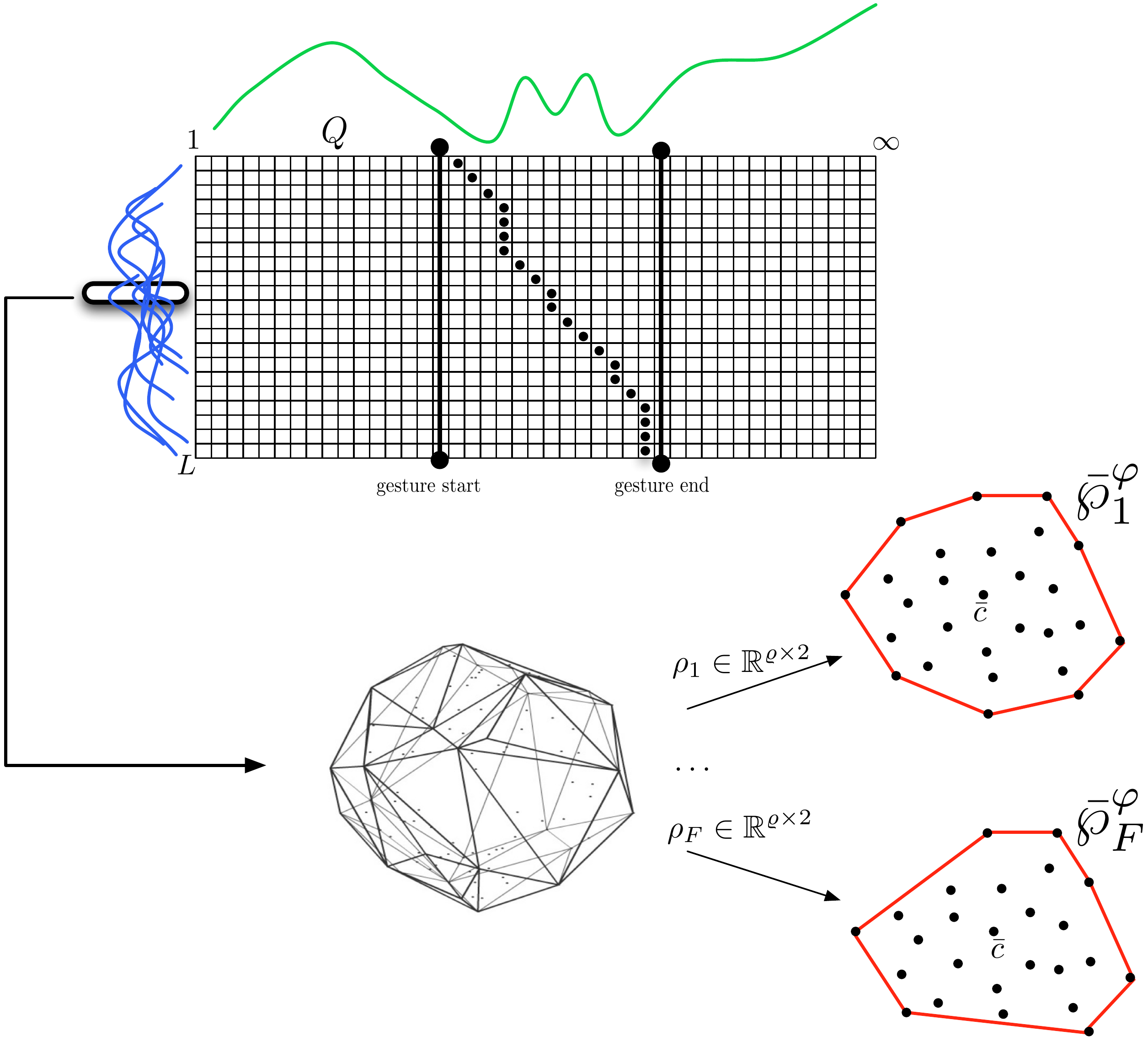}
\end{center}
\caption{Begin-end of gesture recognition of a gesture pattern in
an infinite sequence $Q$ using the probability-based DTW. In this example, APEs are used to model each set of i$-th$ frames.} \label{fig:GMMCH}

\end{figure}
Finally, Algorithm \ref{al:ADTW} shows the proposed DTW algorithm for begin-end gesture detection, where the compute distance $D$ is computed from APE models.
\begin{table}[htbp]\small
\caption{Probability-based DTW applied to begin-end of gesture
recognition, using APEs as base classifiers.}\label{al:ADTW}
\begin{minipage}{8.5 cm}
\begin{description}
\item \textbf{Input:} A gesture model composed by a set of APE models
$C=\{\psi_1,\dots,\psi_m\}$,
a threshold value $\beta$, and the testing sequence
$Q=\{q_1,..,q_{\infty}\}$. Cost matrix $M_{m\times \infty}$ is
defined, where $\mathcal{N}(x), x=(i,j)$ is the set of three upper-left
neighbor locations of $x$ in $M$.
\item \textbf{Output:} Working path $W$ of the detected gesture, if
any.
\item  // Initialization \item
\textbf{for} $i=1 : m$ \textbf{do} \item \quad\quad
\textbf{for} $j=1 : \infty$ \textbf{do}\item
\quad\quad\quad\quad $M(i,j)=\infty$ item
\quad\quad \textbf{end} \vspace{-0.3cm}\item \textbf{end}
\item \textbf{for} $j=1 : \infty$ \textbf{do}
\item \quad\quad $M(0,j)=0$\item
\textbf{end}

\item \textbf{for} $j=0 : \infty$ \textbf{do}
\item \quad\quad \textbf{for} $i=1 : m$ \textbf{do}

\item \quad\quad\quad\quad $x=(i,j)$
\item \quad\quad\quad\quad
$M(x)= \left\lbrace D(x,\psi)+\textrm{min}_{x'\in \mathcal{N}(x)}M(x') \\ \right\rbrace$
\item \quad\quad\textbf{end}

\item \quad\quad \textbf{if} $M(m,t)<\mu$
\textbf{then}

\item \quad\quad\quad\quad
$W=\{\textrm{argmin}_{x'\in \mathcal{N}(x)}M(x')$\}

\item \quad\quad\quad\quad \textbf{return}

\item \quad\quad\textbf{end}
\item\textbf{end}

\end{description}
\end{minipage}
\end{table}

\section{Experimental results}\label{experiments}

In order to present the experimental results, first, we introduce the data, methods, and evaluation measurements of the experiments.

\subsection{The ADHD Behavioural Patterns Dataset}

In this section we introduce the novel dataset in which the experiments are performed. The \textit{ADHD behavioural patterns dataset} is composed of 18 video sequences in which both, a group of three subjects diagnosed with ADHD and three control subjects are recorded in a scholar context, performing recreational and mathematical tasks. These video sequences were recorded using the Kinect$\copyright$ sensor, which is able to obtain RGB and depth information. The features of the dataset are the following:

\begin{itemize}
\item There is an equal proportion of video sequences of ADHD subjects and the control group.
\item There is an equal proportion of video sequences in which the subjects were performing recreational tasks and mathematical tasks.
\item The mean length of the video sequences was approximately 5 minutes each.
\item Outlier events taking place during the recording sessions were manually filtered from the sequences.
\end{itemize}

\begin{figure*}[htbp]
\centering
\begin{tabular}{ccc}
 \includegraphics[width=5.5 cm]{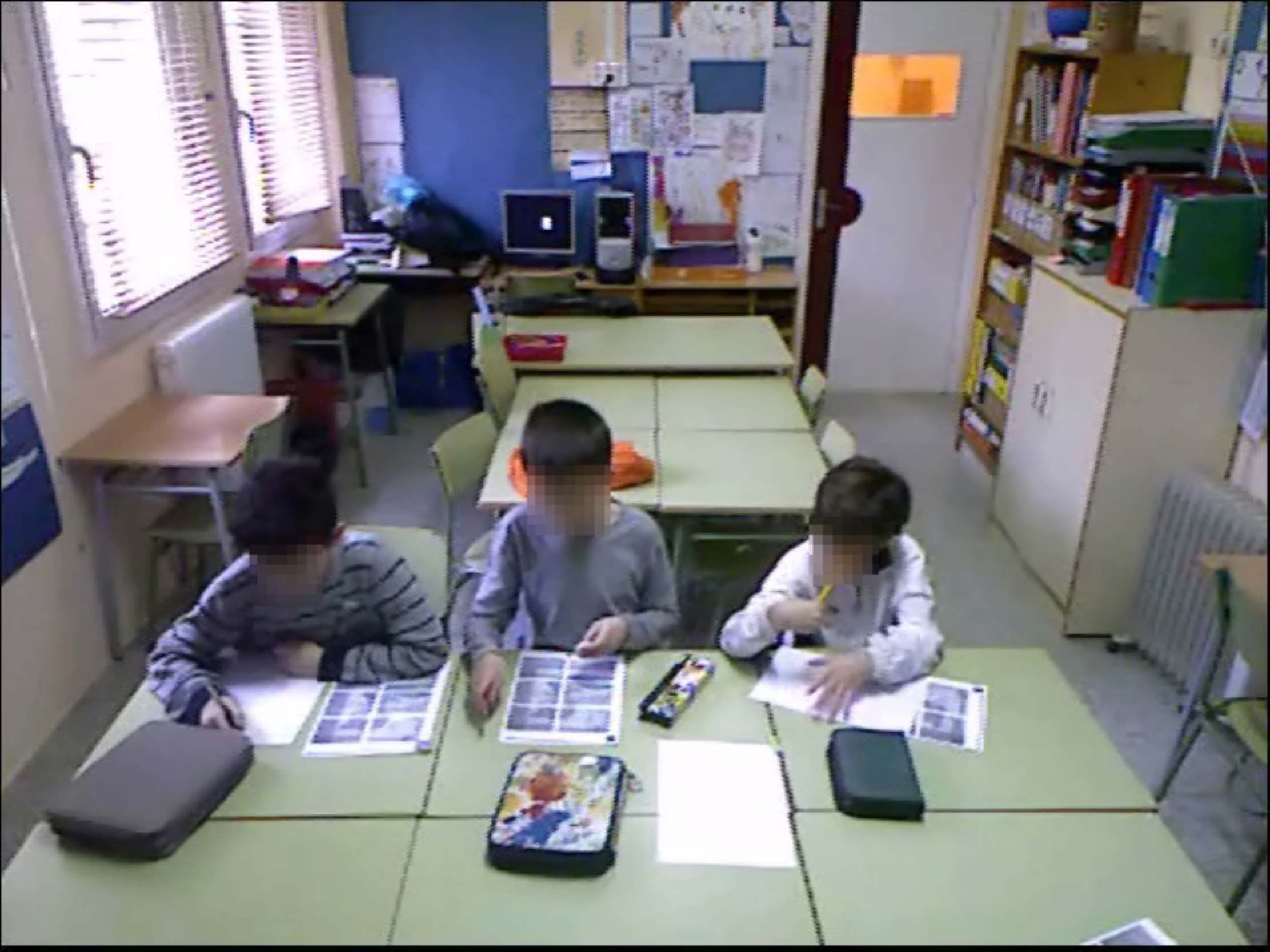} &  \includegraphics[width=5.5 cm]{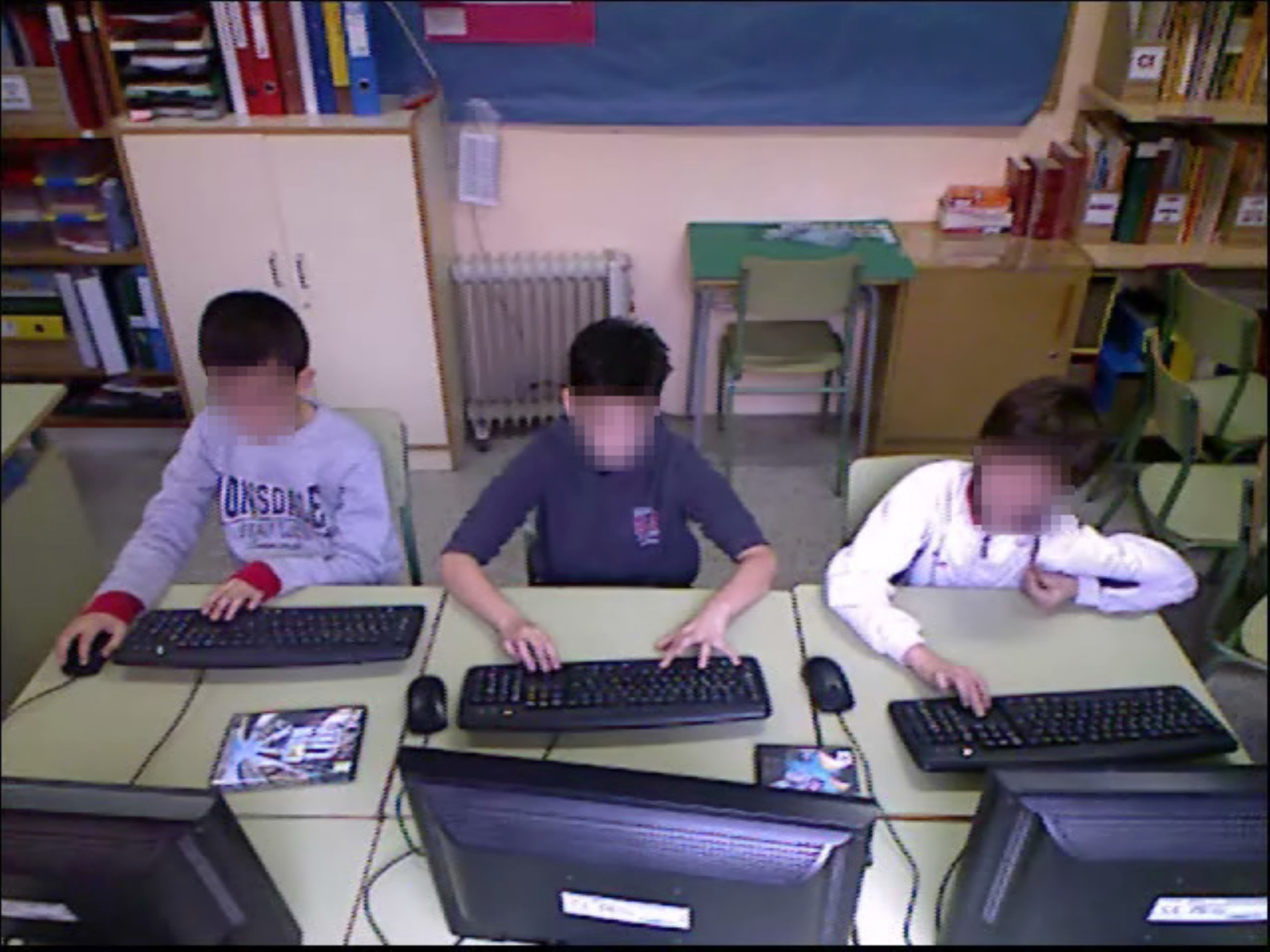} &  \includegraphics[width=5.5 cm]{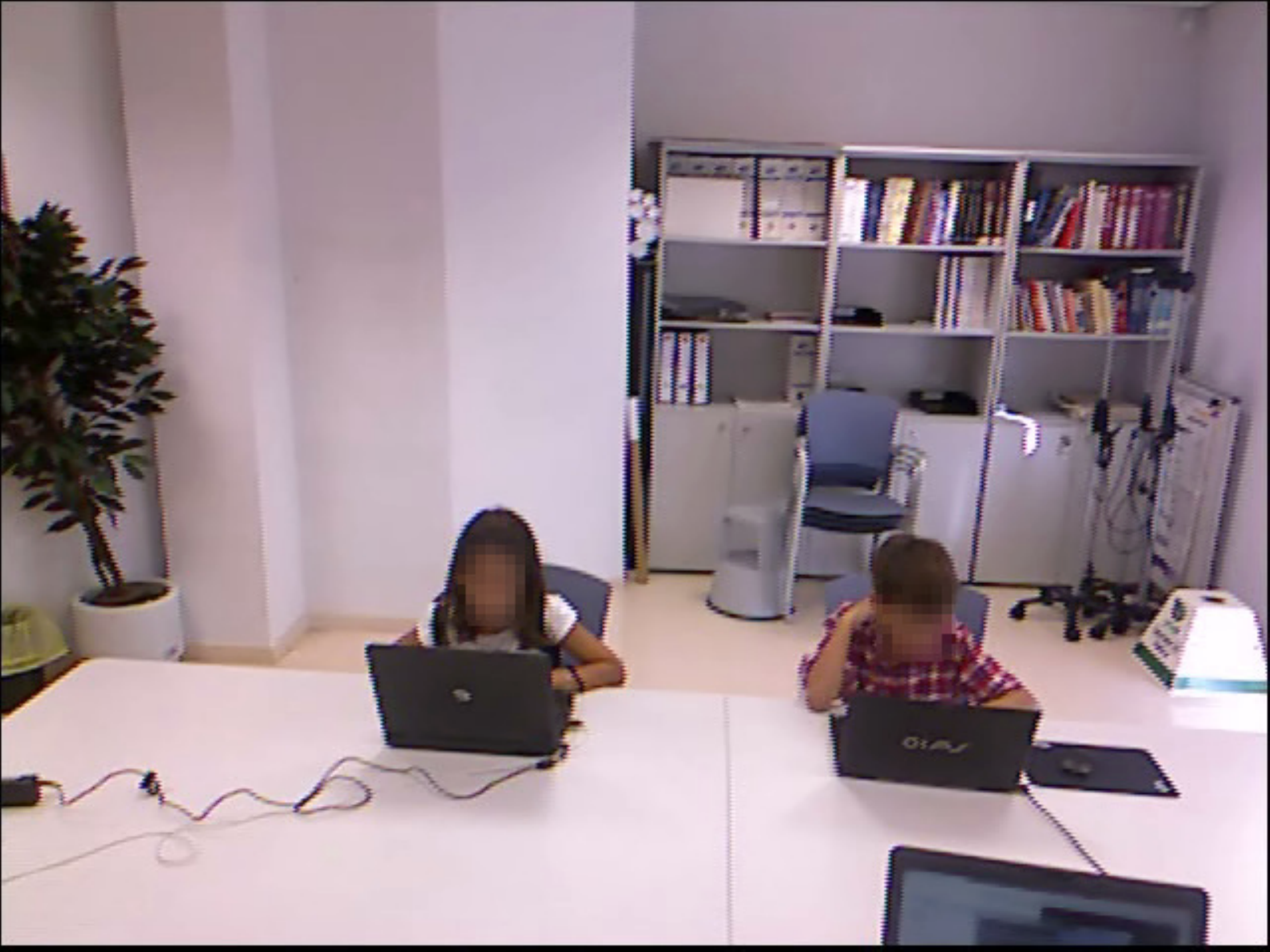} \\
  (a) & (c) & (e) \\
 \includegraphics[width=5.5 cm]{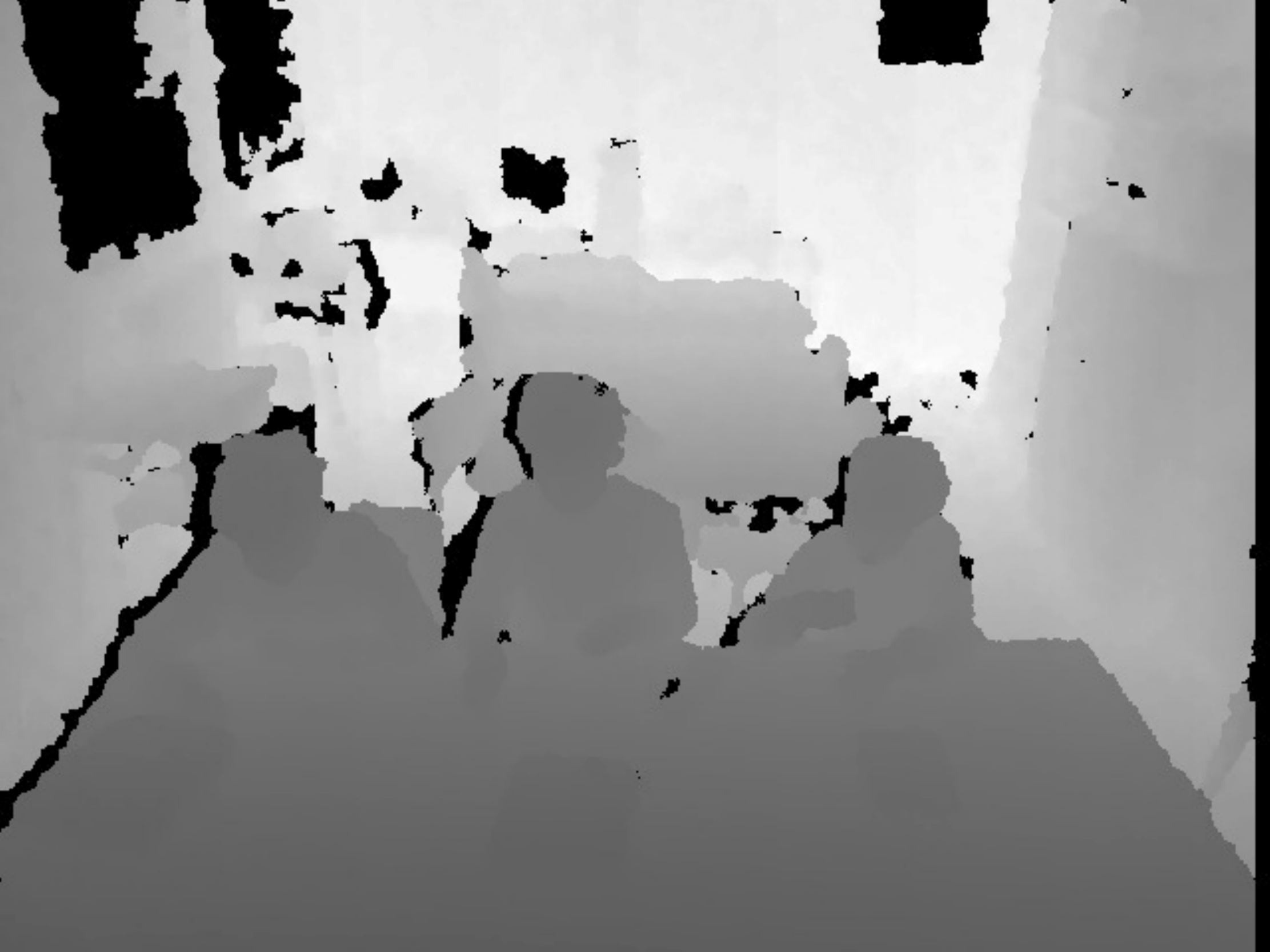} &  \includegraphics[width=5.5 cm]{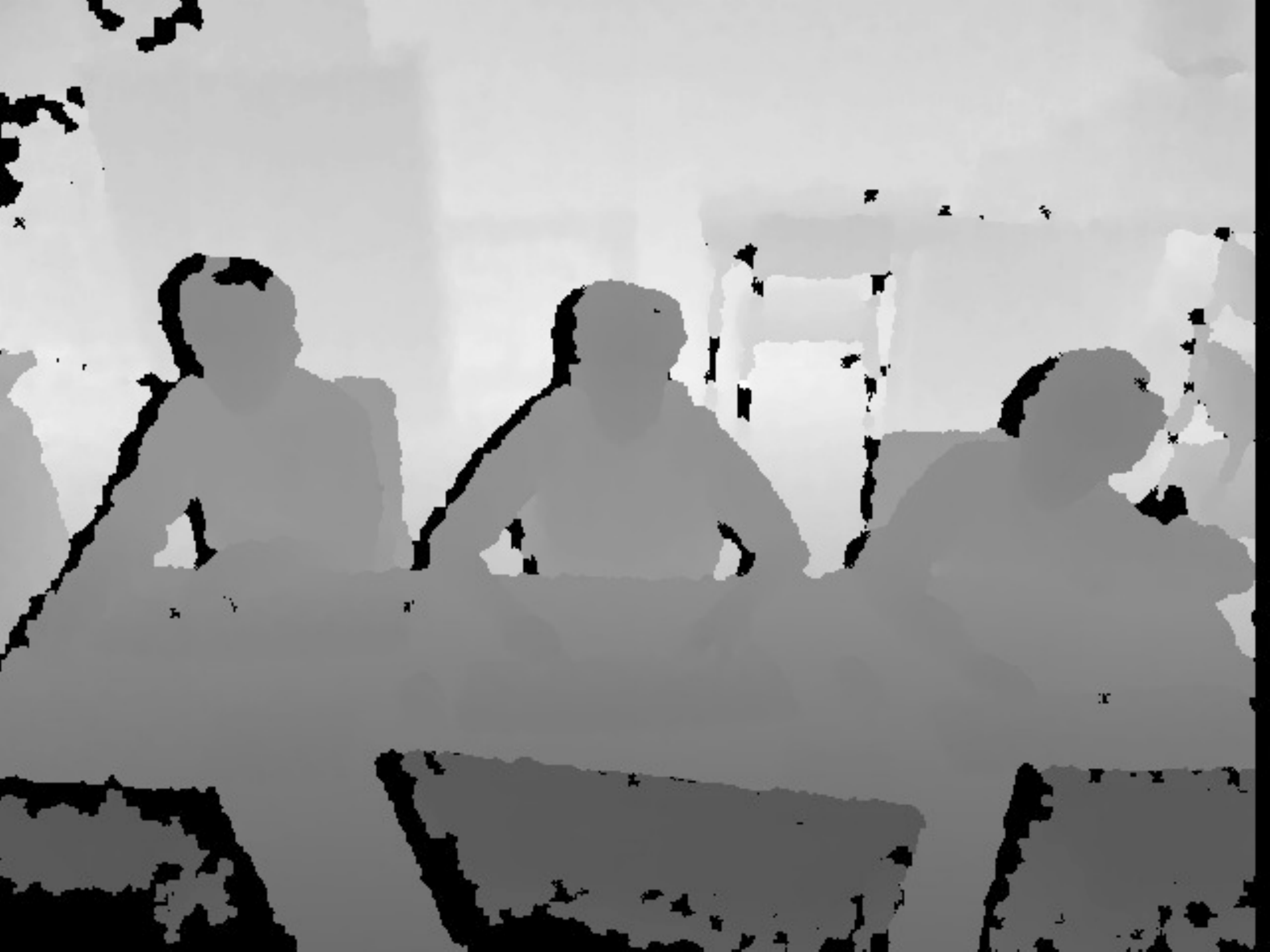} &  \includegraphics[width=5.5 cm]{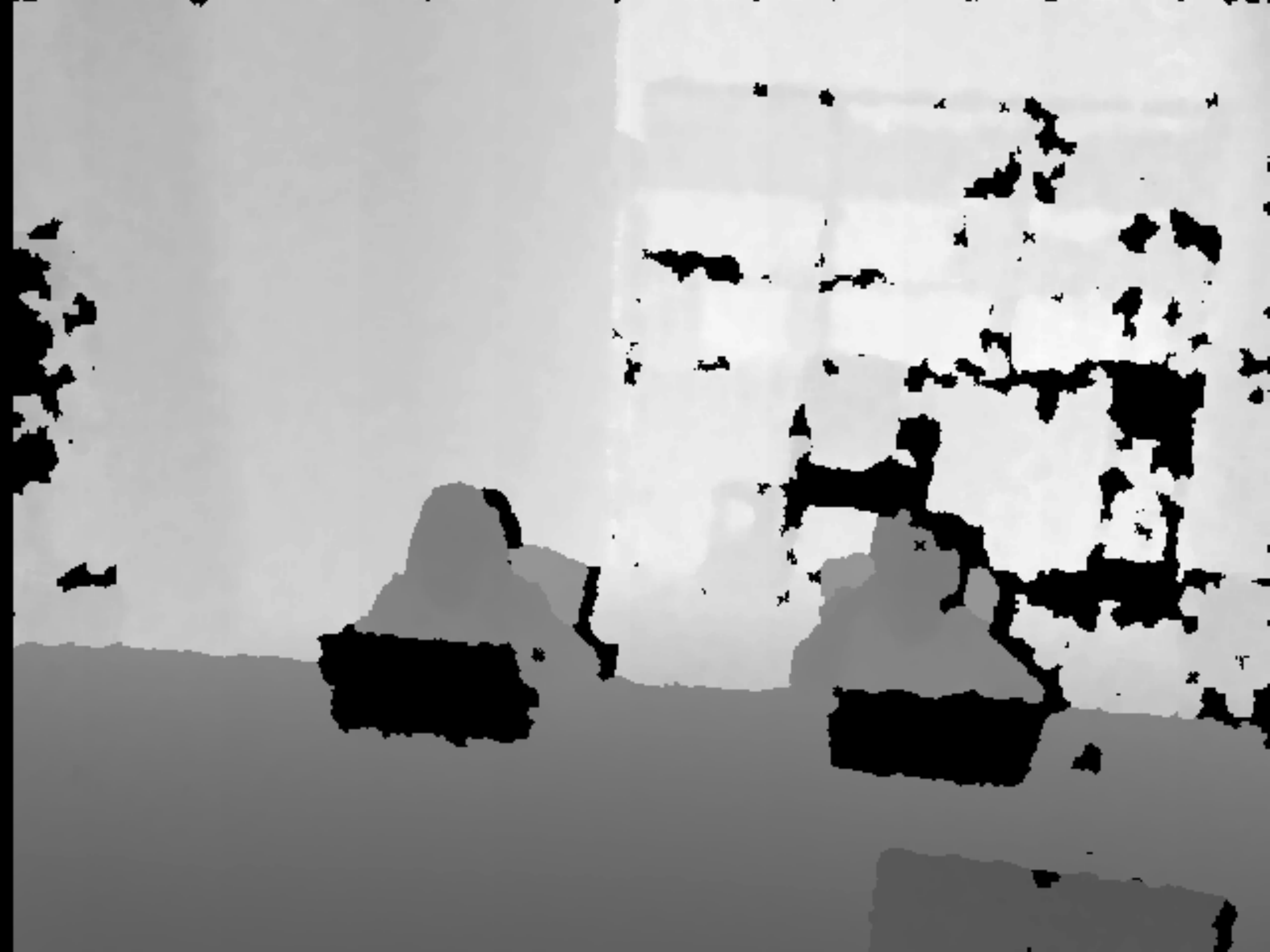} \\ 
   (b) & (d) & (f)  \\
\end{tabular} 
\caption{(a) RGB image of the subjects diagnoses with ADHD performing mathematical tasks. (b) Depth information of ADHD subjects performing mathematical exercises. (c) RGB frame of ADHD subject in the recreational task context. (d) Depth information of ADHD subjects. (e) RGB image of the control group. (f) Depth image of the control group.}
\label{fig::dataset}
\end{figure*}

For each one of the video sequences a manual labelling process was performed, in which two independent observers labelled the start and ending points of each one of the four behavioural patterns defined in Section \ref{indicators} (head turn, torso in table, classmate desk invasion and movement with/without pattern). The agreement of the labelling of the independent observers was measured with the well-known Cohen's kappa coefficient for inter-annotator agreement \cite{cohens}. In order to obtain this measure we used the GSEQ software presented in \cite{gseq}. Finally, the mean Cohen's Kappa statistic of the labelling procedure was $0.93$, which follows in the interval defined as \textit{almost perfect agreement} in \cite{cohens2}, and thus, this labelling is used as the ground truth for evaluating the performance of the proposed methodologies. Table \ref{tab:dataset} shows a summary of the number of samples per subject and behavioural pattern. In addition, in Figure \ref{fig::dataset} some samples of the ADHD behavioural pattern dataset are shown. The dataset is composed of approximately 50.000 frames.

\begin{table}[htbp]
  \centering

  \caption{Number of samples per subject and behavioural patterns.}
    \begin{tabular}{|c|c|c|c|c|c|}
    \hline
& \textbf{Subj. 1} & \textbf{Subj. 2} & \textbf{Subj. 3} & \textbf{Subj. 4} & \textbf{Subj. 5} \\\hline
    \textbf{Head Turn} & 14    & 17     & 24     & 2 &  3\\\hline
    \textbf{Torso in Table} & 4    & 3     & 5     & 0 & 0\\\hline
    \textbf{Class. Inv.} & 7    & 8     & 7     & 0 & 0\\\hline
    \textbf{Movement} & 110     & 98     & 130     & 9 & 6\\\hline
    \textbf{ADHD} & Yes     & Yes     & Yes     & No & No\\\hline
    \end{tabular}%
  \label{tab:dataset}%
\end{table}%

\subsection{Methods}

We compare the following methods, which have been proposed in the paper:

\begin{itemize}
\item \textbf{DTW random}, aligning the streaming sequence $Q$ with a sample selected randomly from the training set of gesture samples for a certain behavioural pattern, using the standard Euclidean distance.
\item \textbf{DTW mean}, aligning the streaming sequence $Q$ with the mean of the set of warped samples $\widetilde{S}$, using also the Euclidean distance.
\item \textbf{DTW GMM}, where the sequence $Q$ is aligned to a whole gesture category by taking into account the probability of a element in $Q$ on the whole GMM, proposed in Section \ref{GMM}.
\item \textbf{DTW APE}, where the sequence $Q$ is aligned to a certain gesture category by modelling the probability of an element in $Q$ as the number of random projections in which the point lies inside a projected Convex Hull, proposed in Section \ref{CH}.
\end{itemize}

\subsection{Evaluation measurements}

The evaluation measurements are overlapping and
accuracy recognition (in
percentage). For the accuracy analysis, we consider that a gesture is correctly detected if
overlapping in the gesture sub-sequence is greater than
60\% (the standard overlapping value~\cite{overlap}). The overlapping measure is defined by  $\frac{g\bigcap p}{g \bigcup p}$, where $g$ is the ground truth and $p$ the prediction. The cost threshold for all methods was obtained by means of a stratified five-fold cross-validation.  In addition, we apply the Friedman and Nemenyi
tests~\cite{validation} in order to look for statistical
significance among the obtained performances.

\begin{figure}[htbp]
\begin{center}
\includegraphics[width=8.85 cm, height=3 cm]{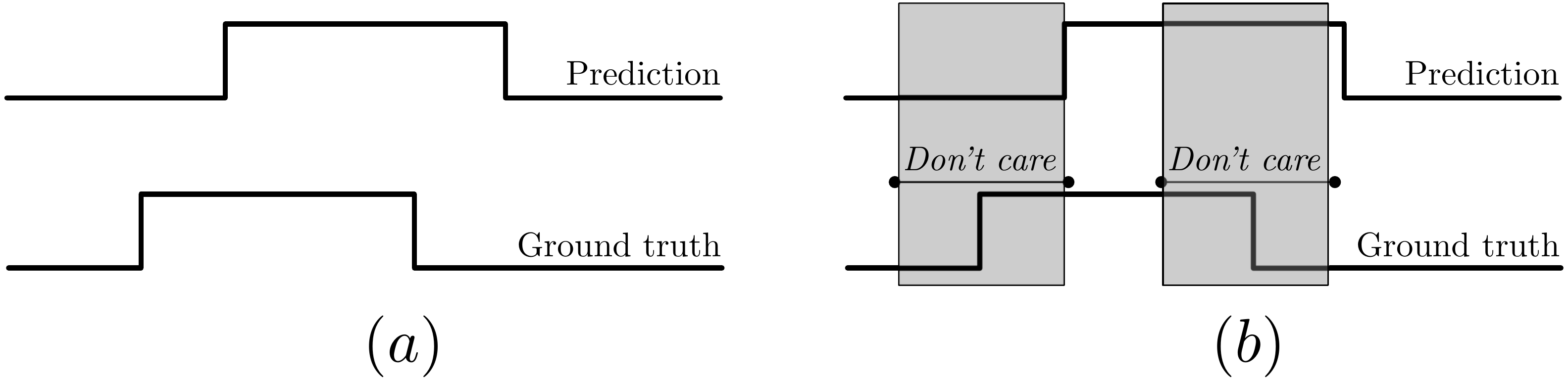}
\end{center}
\caption{ (a) Example of overlapping between a prediction and the ground truth .(b) Example where the \textit{Don't Care} value is used to soften the overlap metric.} \label{fig::dc}
\end{figure}

Furthermore, to allow a deeper analysis of the proposed methodologies and their clinical impact, in our evaluations we use a \textit{'Don't care'} value which provides a more flexible interpretation of the results.
Consider the ground truth of a certain gesture category in a video sequence as a binary vector, which activates when a sample of such category is observed in the sequence. Then, the \textit{'Don't care'} value is defined as the number of bits (frames) which are ignored at the limits of each one of the ground truth instances. Thus, by using this approach we can compensate the pessimistic overlap metric in situations when the detection is shifted some frames. An example of this situation is shown in Figure \ref{fig::dc}.

\subsection{Experimental Results}

\begin{figure*}[htbp]
\setlength{\tabcolsep}{1pt}
\begin{tabular}{cc}
 \includegraphics[width=8.5 cm, height=5 cm]{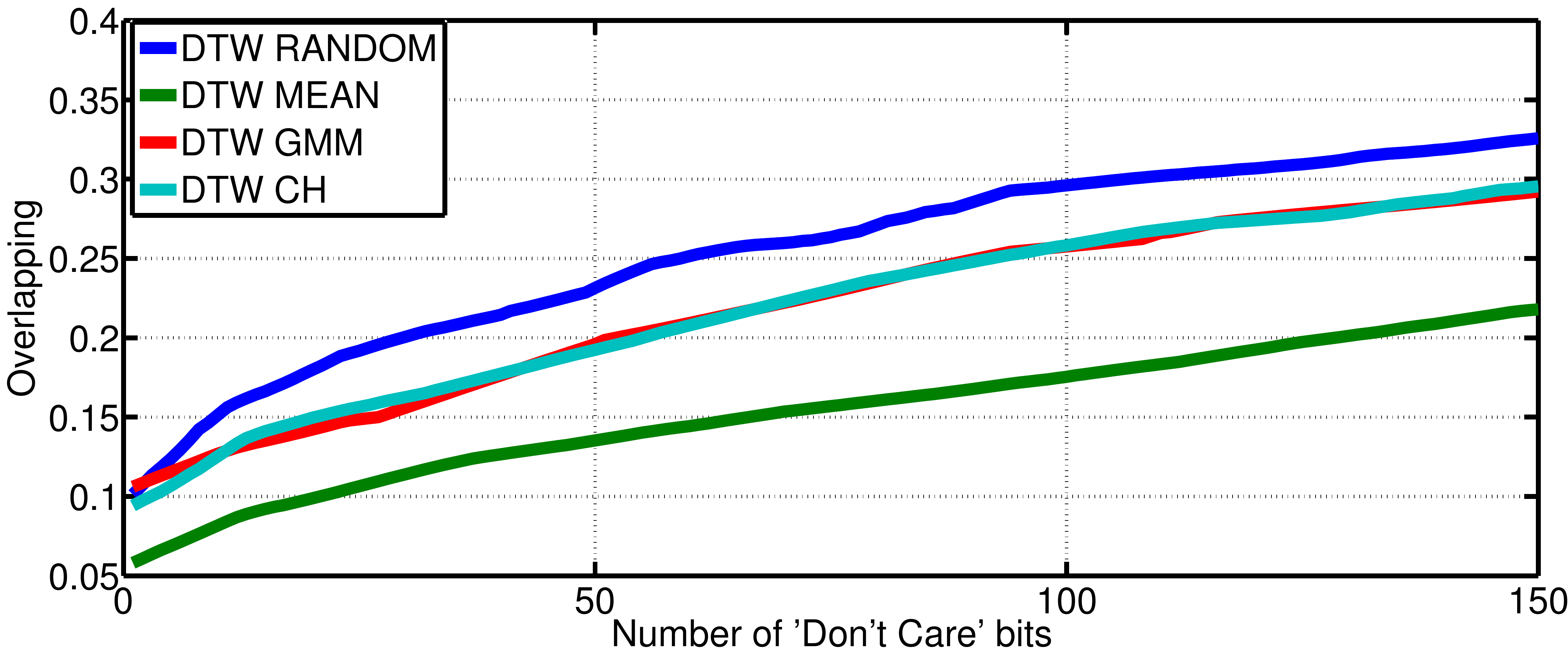} & \includegraphics[width=8.5 cm, height=5 cm]{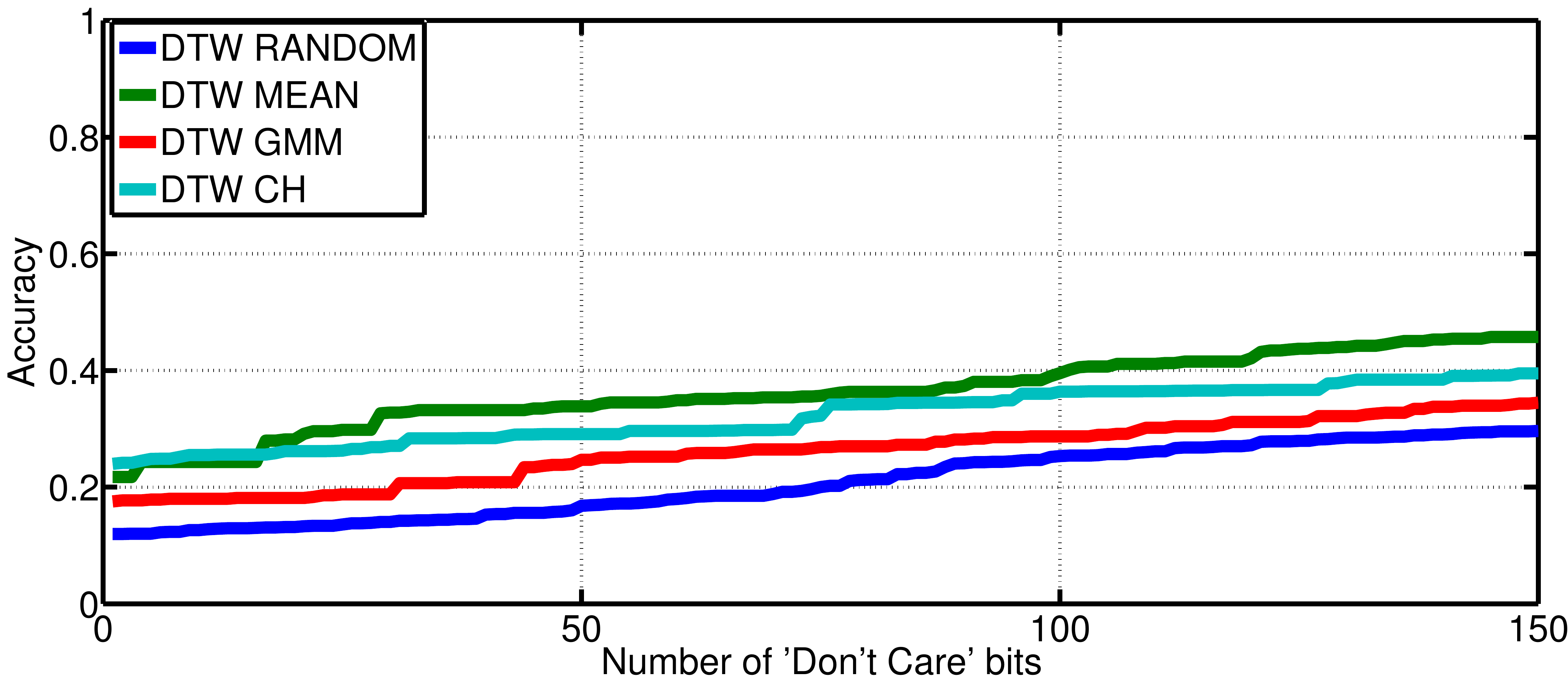}  \\ 
  (a) Head Turn overlapping & (b) Head Turn accuracy\\
 \includegraphics[width=8.5 cm, height=5 cm]{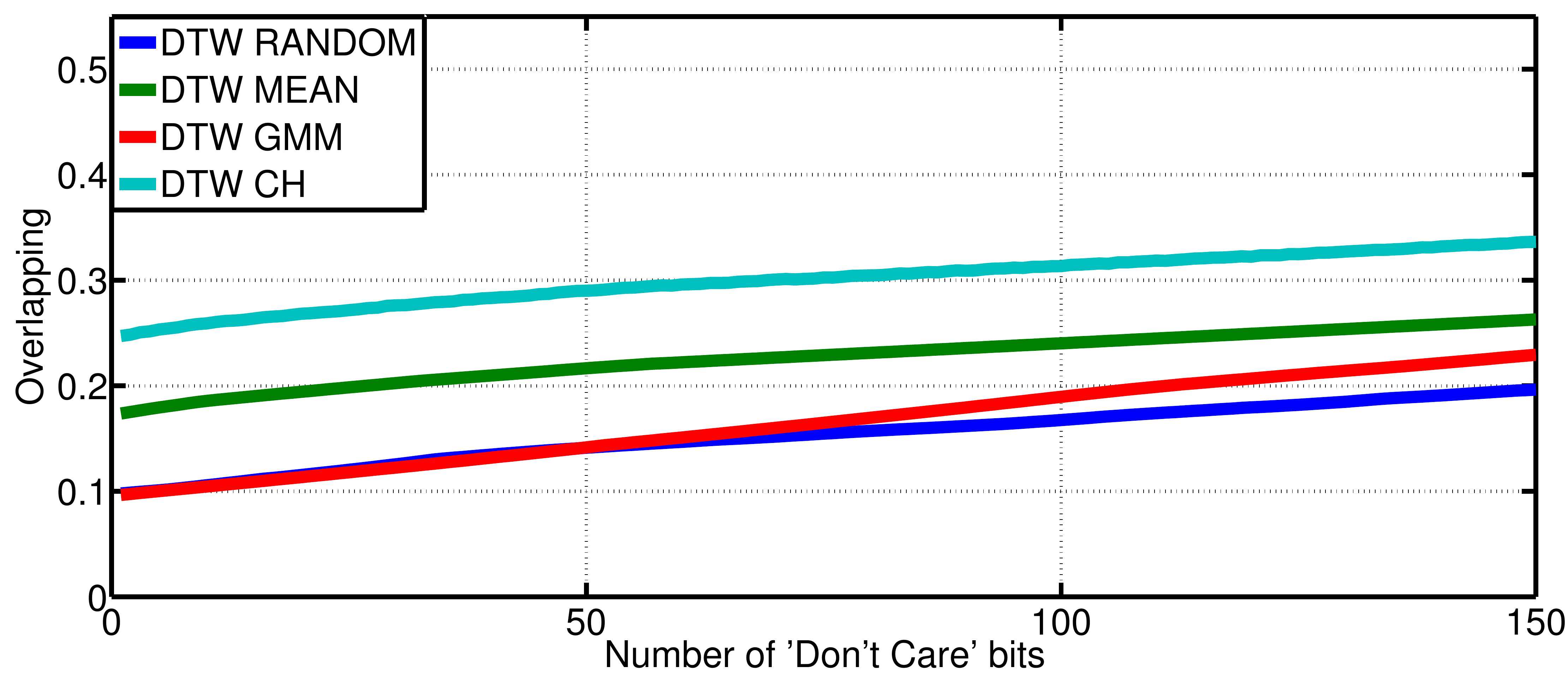} & \includegraphics[width=8.5 cm, height=5 cm]{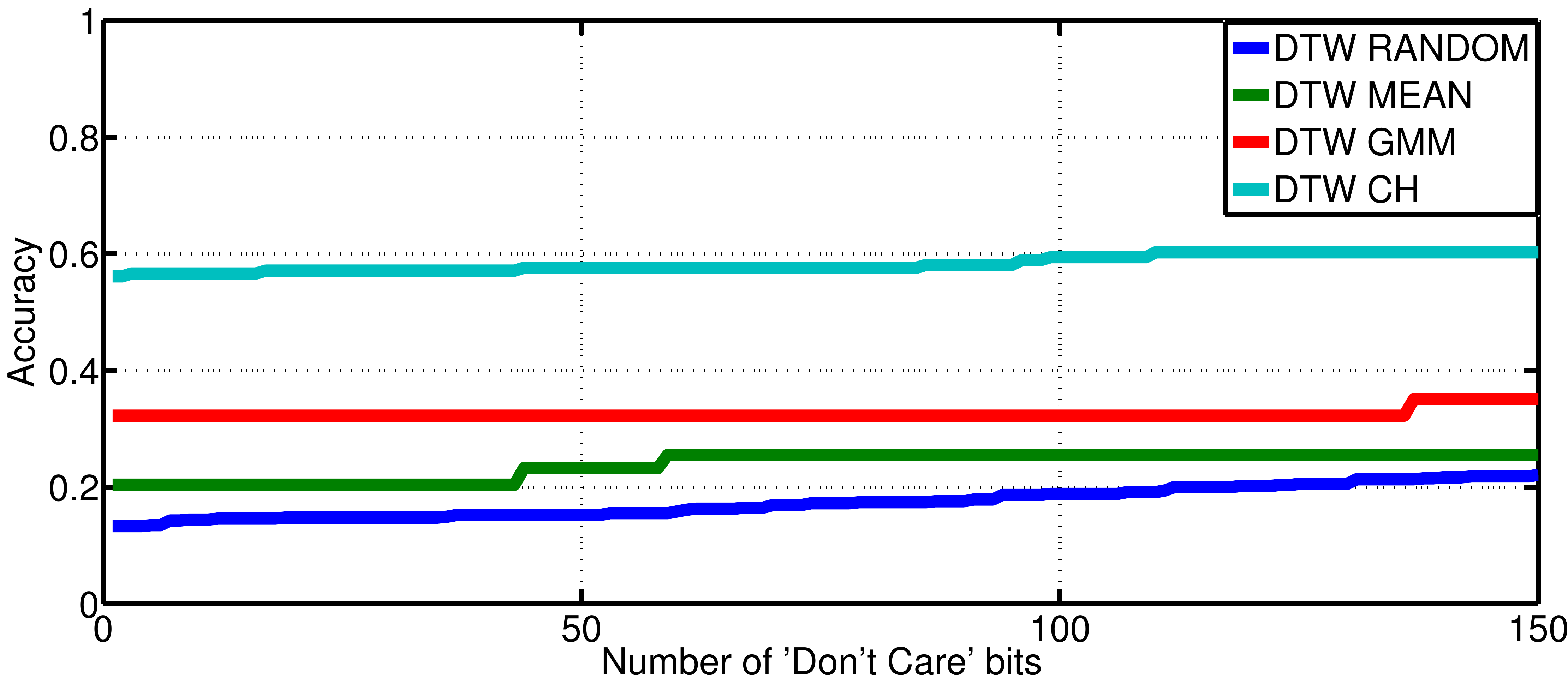}  \\ 
   (c) Torso in Table overlapping & (d) Torso in Table accuracy \\
  \includegraphics[width=8.5 cm, height=5 cm]{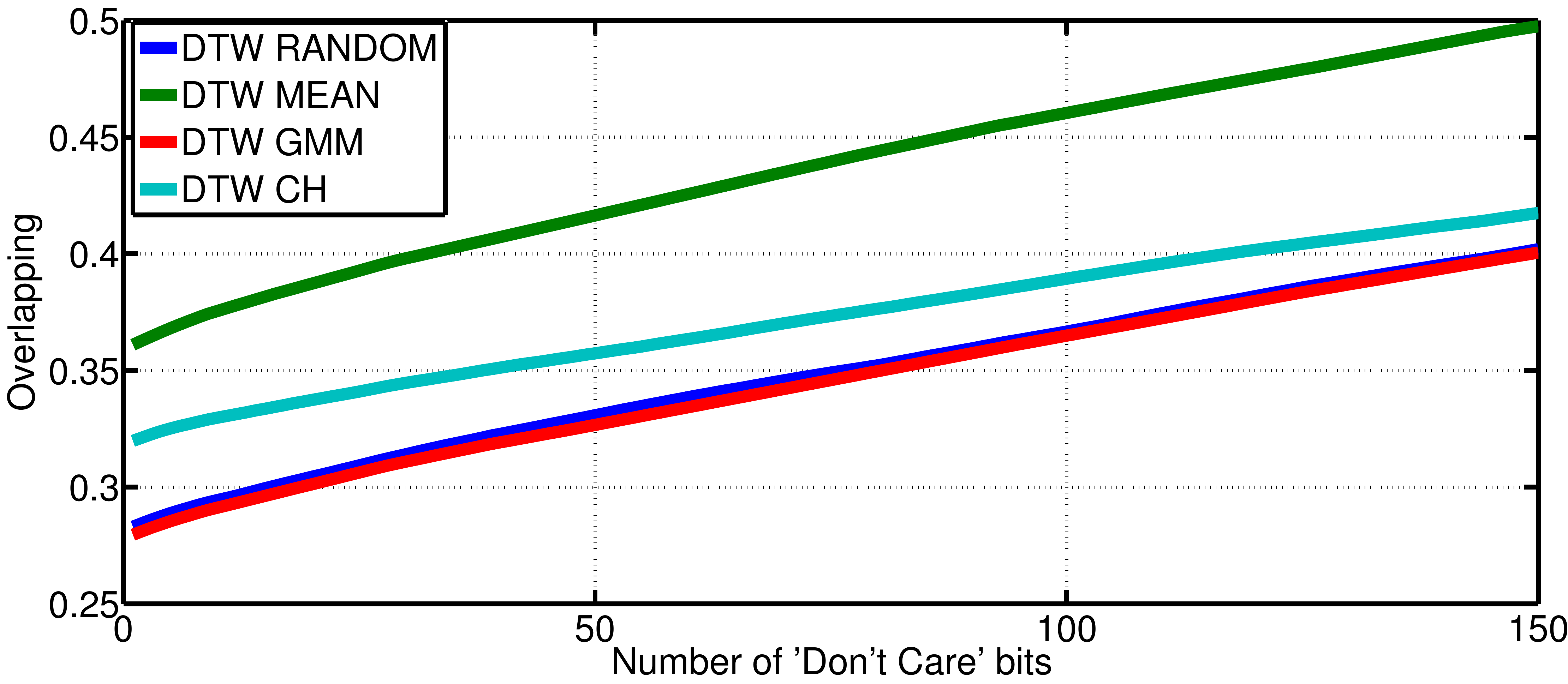} & \includegraphics[width=8.5 cm, height=5 cm]{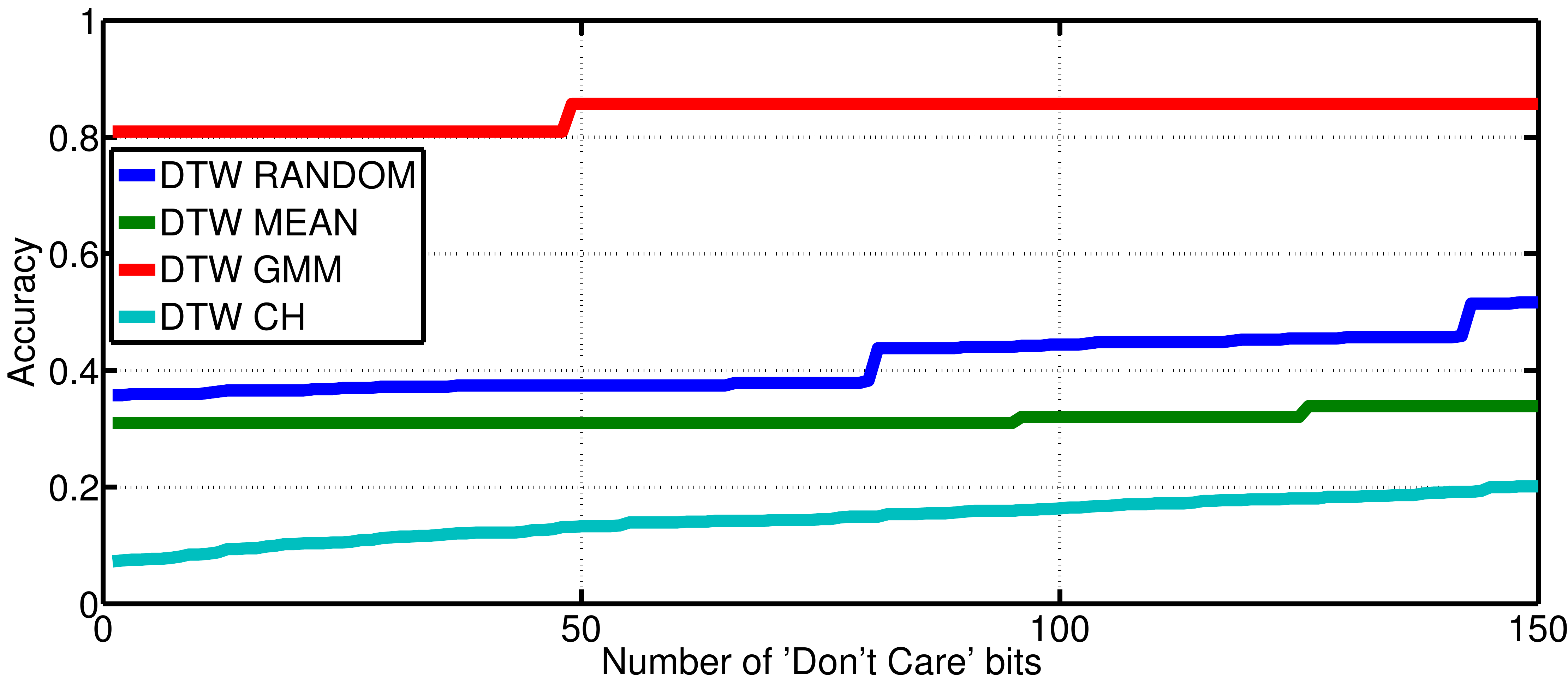}  \\ 
   (e) Classmate Desk Invasion overlapping & (f) Classmate Desk Invasion accuracy\\
  \includegraphics[width=8.5 cm, height=5 cm]{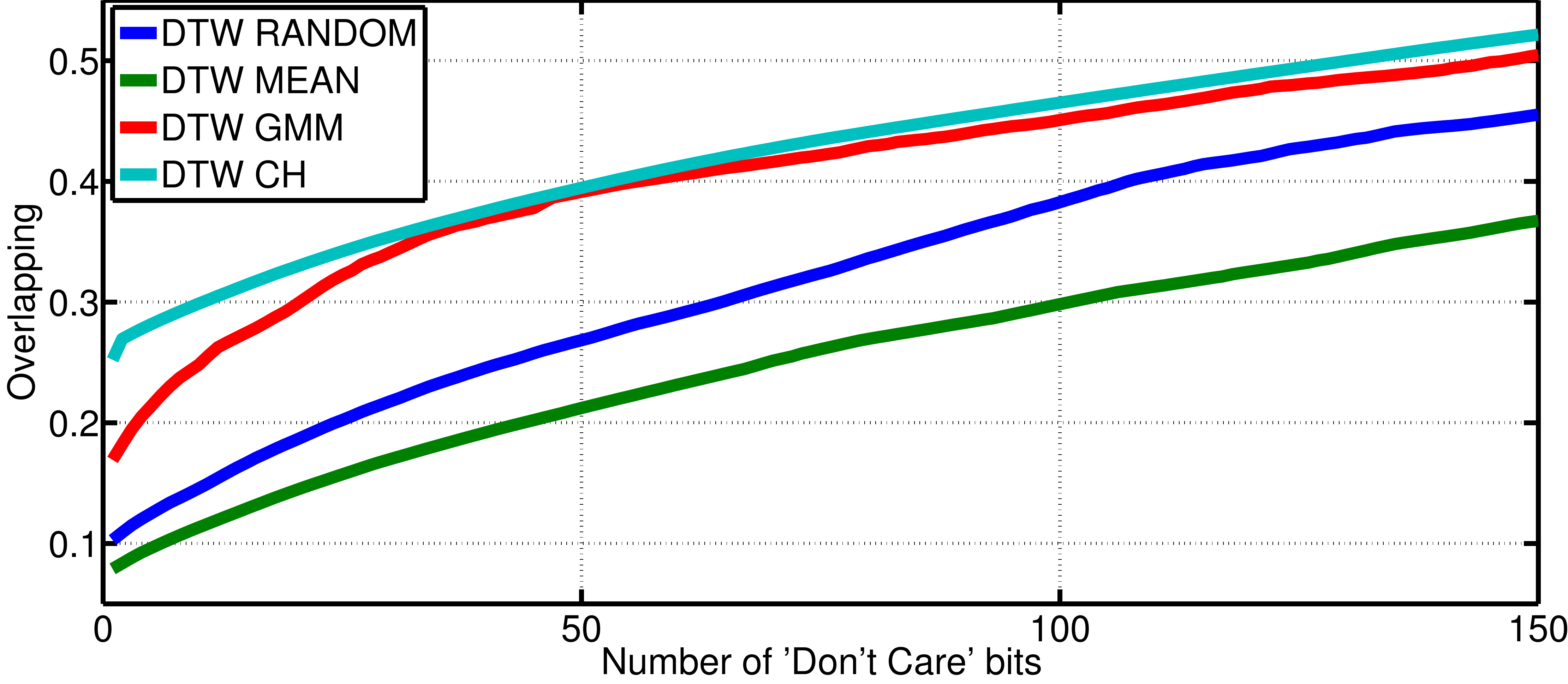} & \includegraphics[width=8.5 cm, height=5 cm]{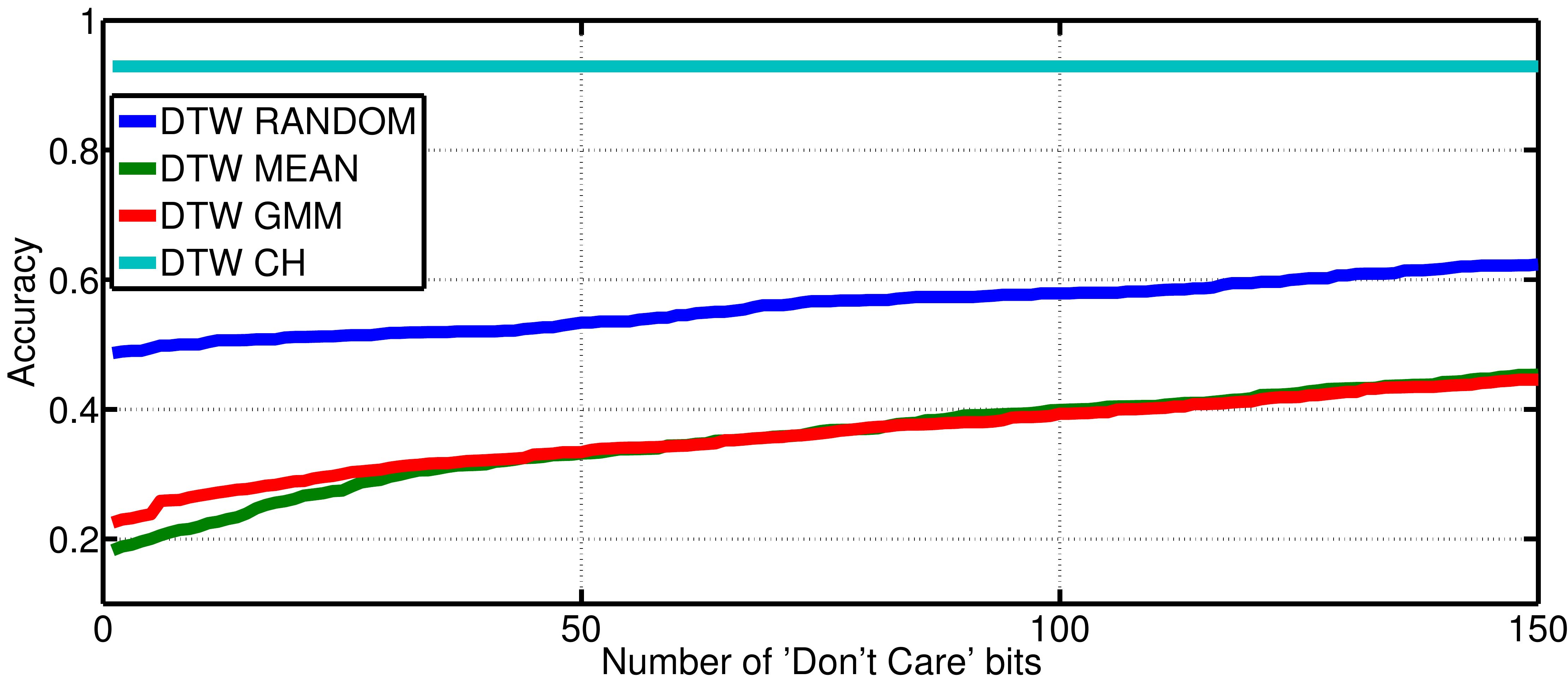}  \\ 
  (g) Movement overlapping & (h) Movement accuracy \\
\end{tabular} 
\caption{(a) Overlapping metric per method and number of \textit{\textit{'Don't Care'}} bits for the \textit{Head Turn} behavioural pattern. (b) Accuracy value for each one of the compared methods and number of \textit{'Don't Care'} bits for the \textit{Head Turn} pattern. (c) Overlapping metric for each method and number of \textit{'Don't Care'} bits for the \textit{Torso in Table} pattern. (d) Accuracy metric per each method and number of \textit{'Don't Care'} bits for the \textit{Torso in Table} behavioural pattern. (e) Overlapping metric and number of \textit{'Don't Care'} bits for the \textit{Classmate Desk Invasion} behavioural pattern. (f) Accuracy value per each compared method and number of \textit{'Don't Care'} bits for the \textit{Classmate Desk Invasion} behavioural pattern. (g) Overlapping metric per method and number of \textit{'Don't Care'} bits for the \textit{Movement} pattern. (h) Accuracy value and number of \textit{'Don't Care'} bits for the \textit{Movement} behavioural pattern.}
\label{fig::ov}
\end{figure*}

Figure \ref{fig::ov} shows the overlapping and accuracy percentages of each one of the compared methods and for each one of the defined behavioural patterns.

In order to present a more reduced and understandable version of the results, we selected specific 'Don't care' values and performed an analysis on those cases. In Tables \ref{table:ov} and \ref{table:ac} we show the overlapping and accuracy values per behavioural pattern and method for certain \textit{\textit{\textit{'Don't Care'}}} values.

% Table generated by Excel2LaTeX from sheet 'Sheet2'
\begin{table}[htbp]
  \centering
  \scriptsize
  \caption{Performance of the compared methodologies in terms of overlapping.}
    \begin{tabular}{|c|c|c|c|c|}
\hline
    \textbf{\textit{Head Turn}} & \textbf{DTW Random} & \textbf{DTW Mean} & \textbf{DTW GMM} & \textbf{DTW CH} \\ \hline

    \textbf{DC 1}  & 0.1012 & 0.0581 & \textbf{0.1015} & 0.0942 \\\hline
    \textbf{DC 50} & \textbf{0.2314} & 0.1352 & 0.1998 & 0.1924 \\\hline
    \textbf{DC 100} & \textbf{0.2960} & 0.1753 & 0.2673 & 0.2582 \\\hline
    \textbf{DC 150} & \textbf{0.3257} & 0.2179 & 0.3096 & 0.2954\\\hline
\hline\hline
      \textbf{\textit{Torso in Table}} & \textbf{DTW Random} & \textbf{DTW Mean} & \textbf{DTW GMM} & \textbf{DTW CH} \\\hline
    \textbf{DC 1}  & 0.0979 & 0.1737 & 0.0966 & \textbf{0.2521} \\\hline
    \textbf{DC 50} & 0.1412 & 0.2165 & 0.1415 & \textbf{0.2901} \\\hline
    \textbf{DC 100} & 0.1675 & 0.2402 & 0.1895 & \textbf{0.3134} \\\hline
    \textbf{DC 150} & 0.1964 & 0.2628 & 0.2293 & \textbf{0.3364} \\\hline
\hline\hline
      \textbf{\textit{Classmate Inv.}} & \textbf{DTW Random} & \textbf{DTW Mean} & \textbf{DTW GMM} & \textbf{DTW CH} \\\hline
    \textbf{DC 1}  & 0.2830 & \textbf{0.3610} & 0.2796 & 0.3198 \\\hline
    \textbf{DC 50} & 0.3308 & \textbf{0.4164} & 0.3266 & 0.3573 \\\hline
    \textbf{DC 100} & 0.3666 & \textbf{0.4603} & 0.3649 & 0.3893 \\\hline
    \textbf{DC 150} & 0.4019 & \textbf{0.4975} & 0.4005 & 0.4174 \\\hline
\hline\hline
      \textbf{\textit{Movement}} & \textbf{DTW Random} & \textbf{DTW Mean} & \textbf{DTW GMM} & \textbf{DTW CH} \\\hline
    \textbf{DC 1 } & 0.1028 & 0.0789 & 0.1682 & \textbf{0.2521} \\\hline
    \textbf{DC 50} & 0.2683 & 0.2121 & 0.3718 & \textbf{0.3945} \\\hline
    \textbf{DC 100} & 0.3826 & 0.2981 & 0.4429 & \textbf{0.4651} \\\hline
    \textbf{DC 150} & 0.4551& 0.3672 & 0.5044 & \textbf{0.5215} \\\hline

    \end{tabular}%
  \label{table:ov}%
\end{table}%

% Table generated by Excel2LaTeX from sheet 'Sheet2'
\begin{table}[htbp]
  \centering
  \scriptsize
  \caption{Performance of the compared methodologies based on the accuracy metric.}
    \begin{tabular}{|c|c|c|c|c|}
    \hline
    \textit{\textbf{Head Turn}} & \textbf{DTW Random} & \textbf{DTW Mean} & \textbf{DTW GMM} & \textbf{DTW CH} \\\hline
    \textbf{DC 1} & 0.1198 & 0.2174 & 0.1755 & \textbf{0.2399} \\\hline
    \textbf{DC 50} & 0.168 & \textbf{0.3386} & 0.2469 & 0.2909 \\\hline
    \textbf{DC 100} & 0.2534 & \textbf{0.3955} & 0.2871 & 0.3638 \\\hline
    \textbf{DC 150} & 0.2963 & \textbf{0.4573} & 0.3448 & 0.3951 \\\hline
\hline\hline
    \textit{\textbf{Torso in Table}} & \textbf{DTW Random} & \textbf{DTW Mean} & \textbf{DTW GMM} & \textbf{DTW CH} \\\hline
    \textbf{DC 1} & 0.1331 & 0.2043 & 0.3227 & \textbf{0.5614} \\\hline
    \textbf{DC 50} & 0.1525 & 0.2328 & 0.3227 & \textbf{0.5760} \\\hline
    \textbf{DC 100} & 0.1885 & 0.2551 & 0.3227 & \textbf{0.5941} \\\hline
    \textbf{DC 150} & 0.2215 & 0.2551 & 0.3512 & \textbf{0.6024} \\\hline
\hline\hline
    \textit{\textbf{Classmate Inv.}} & \textbf{DTW Random} & \textbf{DTW Mean} & \textbf{DTW GMM} & \textbf{DTW CH} \\\hline
    \textbf{DC 1} & 0.3575 & 0.3100  & \textbf{0.8096} & 0.0729 \\\hline
    \textbf{DC 50} & 0.3743 & 0.3100 & \textbf{0.8572} & 0.1330 \\\hline
    \textbf{DC 100} & 0.4445 & 0.3204 & \textbf{0.8572} & 0.1636 \\\hline
    \textbf{DC 150} & 0.5168 & 0.3389 & \textbf{0.8572} & 0.2015 \\\hline
\hline\hline
    \textit{\textbf{Movement}} & \textbf{DTW Random} & \textbf{DTW Mean} & \textbf{DTW GMM} & \textbf{DTW CH} \\\hline
    \textbf{DC 1} & 0.1827 & 0.2254 & 0.4870 & \textbf{0.9291} \\\hline
    \textbf{DC 50} & 0.3321 & 0.3340 & 0.5339 & \textbf{0.9291} \\\hline
    \textbf{DC 100} & 0.3992 & 0.3933 & 0.5789 & \textbf{0.9291} \\\hline
    \textbf{DC 150} & 0.4536 & 0.4458 & 0.6238 & \textbf{0.9291} \\\hline
    \end{tabular}%
  \label{table:ac}%
\end{table}%

Finally, Table \ref{tab:total} shows the mean rank per each methodology and the final mean rank.

% Table generated by Excel2LaTeX from sheet 'Sheet2'
\begin{table}[htbp]
  \centering
  \scriptsize
  \caption{Mean ranks for each method and certain 'Don't care values'.}
    \begin{tabular}{|c|c|c|c|c|}
\hline
    \textit{\textbf{Mean rank}} & \textbf{DTW Random} & \textbf{DTW Mean} & \textbf{DTW GMM} & \textbf{DTW CH} \\\hline
    \textbf{DC 1} & 3.1250 & 2.7500  & 2.3750 & \textbf{1.7500} \\\hline
    \textbf{DC 50} & 3.1250 & 2.625 & 2.3750 &\textbf{ 1.8750} \\\hline
    \textbf{DC 100} & 3.0000    & 2.7500  & 2.3750 & \textbf{1.8750} \\\hline
    \textbf{DC 150} & 3.0000    & 2.7500  & 2.3750 & \textbf{1.8750} \\\hline
    \textbf{Overall mean} & 3.0625 & 2.7187 & 2.3750 & \textbf{1.8437} \\\hline
    \end{tabular}%
  \label{tab:total}%
\end{table}%

Once all the rankings are computed, in order to reject the null hypothesis that the measured performance ranks differ from
the mean performance rank, and that the performance ranks are affected by randomness in
the results, we use the Friedman test. Thus, with $h=4$ methods to compare and $U=$ $4$ behavioural patterns $\times$ 4 \textit{Don't care} values (1,50,100,150) $\times$ 2 metrics (overlapping and accuracy) $= 32$,  the Friedman statistic value is computed as follows, where $V$ is the mean rank:

\begin{equation}
X_F^2 = \frac{12U}{h(h+1)}\left[\sum_jV_j^2
-\frac{h(h+1)^2}{4}\right]. \label{friedman}
\end{equation}

In our case, with $h=4$ DTW methods to compare, $X_F^2=14.8875$.
Since this value is undesirable conservative, Iman and Davenport
proposed a corrected statistic:
\begin{equation}
F_F=\frac{(U-1)X_F^2}{U(h-1)-X_F^2}. \label{friedman2}
\end{equation}

Applying this correction we obtain $F_F=5.68 $. With four methods
and 32 experiments, $F_F$ is distributed according to the $F$
distribution with 3 and 91 degrees of freedom. The critical value
of $F(3,93)$ for $0.05$ is $0.12$. As the value of $F_F$ is 
higher than $0.12$ we can reject the null hypothesis.

Furthermore, we perform a Nemenyi test in order to check if any of these methods can be singled out \cite{validation}, the Nemenyi statistic is obtained as follows:

\begin{equation}
CD=q_{\alpha}\sqrt{\frac{h(h+1)}{6U}}.\label{nemenyi}
\end{equation}

\begin{figure}[htbp]
\begin{center}
\includegraphics[width=8.85 cm, height=2.5 cm]{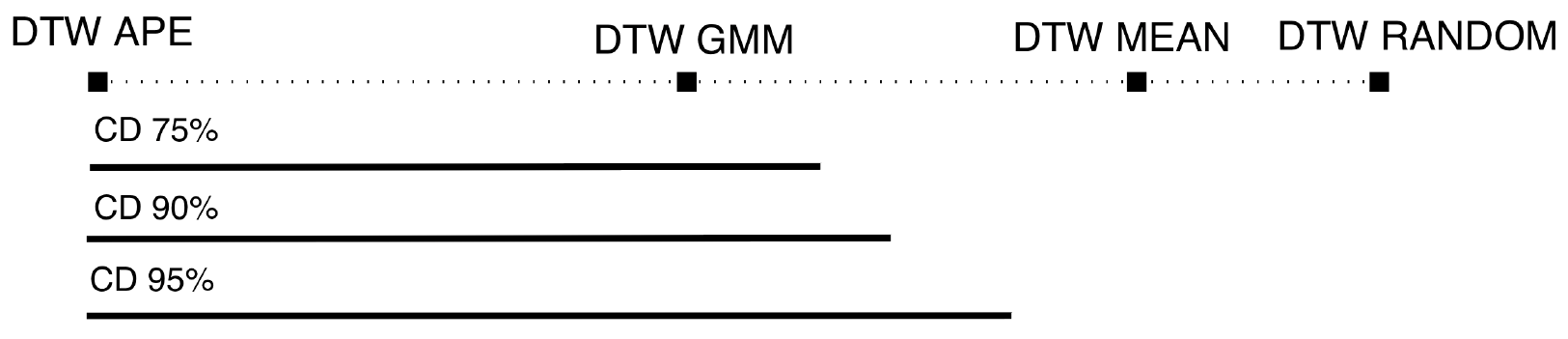}
\end{center}
\caption{ Mean rank and confidence interval per method.} \label{fig::cd}
\end{figure}

In our case, for $k=4$ DTW methods to compare and $N=32$ experiments the critical value for a $95\%$ of confidence is $CD_{0.95} = 2.569 \cdot \sqrt{\frac{20}{192}} = 0.8291 $.  As a result non of the standard DTW methods intersect with our proposal of DTW GMM or DTW CH which is the best in mean ranking. This results are highly desirable since they supports the fact that the \textit{proposed methodologies obtain a statistically significant improvement} in performance when compared to standard DTW approaches. For completion, we also compute the $CD_{0.90}$ and $CD_{0.75}$; results are shown in Figure \ref{fig::cd}.

These results support the fact that our proposal DTW APE is statistically better than the standard DTW approaches, obtaining very satisfying results while keeping similar computational complexity. In addition, though our contribution can be applied to any general purpose gesture recognition system, from a clinical point of view, the presented analyses were reported as relevant by physicians involved in the project and specialists on ADHD from hospitals in the area of Catalonia.

\section{Conclusions and Future work}\label{conclusions}

In this paper we presented an extension of the DTW algorithm in order to handle the intra-class variability of a gesture class. This variability was encoded using one-class classifiers, such as, GMMs and APEs. In order to be able to embed these classifiers in the DTW context, the association cost was redefined to take into account the properties of such classifiers. We applied this extension in a real world problem, detecting ADHD behavioural patterns to support clinicians in diagnose purposes. In our experiments, on a novel multi-modal ADHD dataset, the proposed methodology obtained statistically significant improvements with respect to DTW techniques while obtaining relevant classification rates from a clinical point of view.

The results of this study motivate the use of the proposed techniques with a much broader set of ADHD behavioural patterns in order to provide additional information to the clinician. Moreover, the presented methodology represents a significant contribution for general purpose Human Behaviour Analysis systems.

\section*{Acknowledgements}
This work is partly supported by projects IMSERSO-Ministerio de Sanidad 2011 Ref. MEDIMINDER and RECERCAIXA 2011 Ref. REMEDI, SUR-DEC of the Generalitat de Catalunya and FSE, and TIN2013-43478-P. The work of Antonio is supported by an FPU fellowship from the Ministerio de Educacion of Spain.

\bibliographystyle{abbrv}      % basic style, author-year citations
\bibliography{latex12}   % name your BibTeX data base

\begin{IEEEbiography}[{\includegraphics[width=1in,height=1.25in,clip,keepaspectratio]{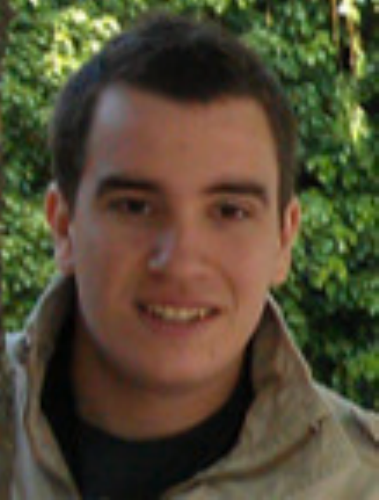}}]{Miguel \'{A}ngel Bautista}
 received his B. Sc. and M. Sc. degrees in Computer Science and Artificial Intelligence from Universitat de Barcelona and Universitat Politécnica de Catalunya  respectively in 2010. He is a research member at Computer Vision Center at Universitat Autonoma de Barcelona,  Applied Math and Analysis Dept. at Universitat de Barcelona and BCN Perceptual Computing Lab and Human Pose Recovery and Behavior Analysis Group at University of Barcelona . In 2010 Miguel Angel received the first prize from the Catalan Association of Artificial Intelligence Thesis Awards. Currently Miguel Angel is pursuing a Ph. D in Error Correcting Output Codes as a theoretical framework to treat multi-class and multi-label problems. His interests are, between others, Machine Learning, Computer Vision, Convex Optimization and its applications into Human Gesture analysis.
\end{IEEEbiography}
\vspace{-1 cm}
\begin{IEEEbiography}[{\includegraphics[width=1in,height=1.25in,clip,keepaspectratio]{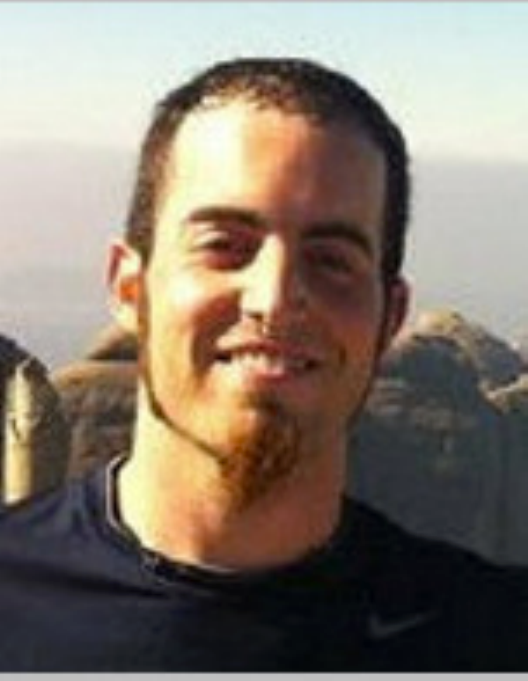}}]{Antonio Hern\'{a}ndez-Vela}
 received his Bachelor degree in Computer Science and M.S. degree in Computer Vision and Artificial Intelligence at Universitat Autònoma de Barcelona (UAB) in 2009 and 2010, respectively. He is currently a research member at the Computer Vision Center (UAB) and PhD student at University of Barcelona. He is also member of the BCN Perceptual Computing Lab research group and Human Pose Recovery and Behavior Analysis Group. He is mainly interested in the application of Computer Vision and Artificial Intelligence techniques to projects that can help impaired people to improve their life quality, especially in the area of human pose recovery and behaviour analysis. 
\end{IEEEbiography}
\vspace{-1 cm}
\begin{IEEEbiography}[{\includegraphics[width=1in,height=1.25in,clip,keepaspectratio]{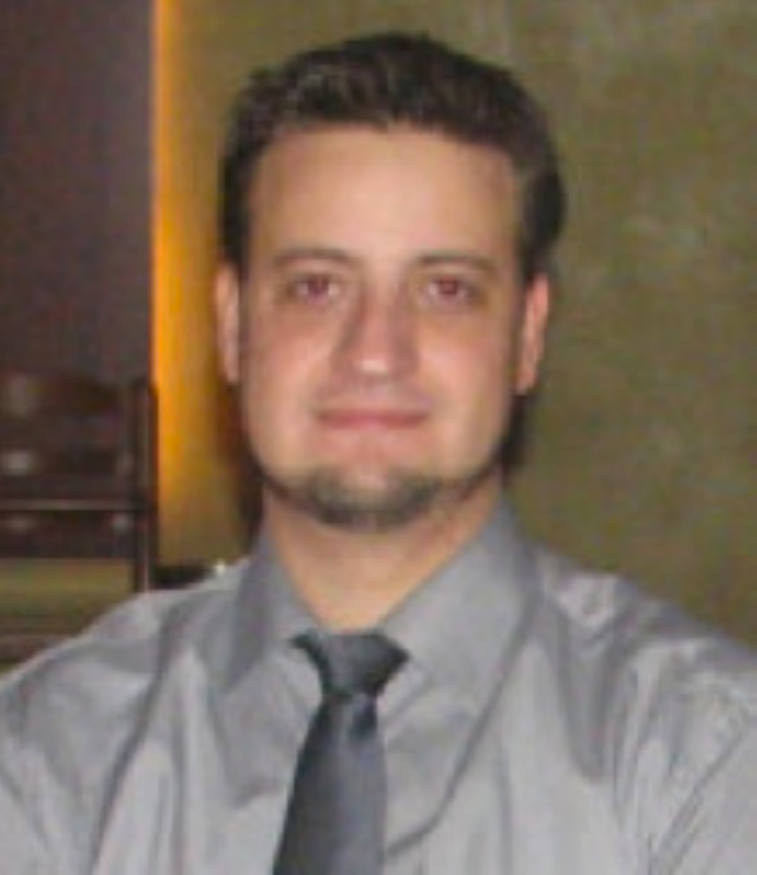}}]{Sergio Escalera}
Sergio Escalera received the B.S. and M.S. degrees from the Universitat Autònoma de Barcelona (UAB), Barcelona, Spain, in 2003 and 2005, respectively. He obtained the P.h.D. degree on Multi-class visual categorization systems at Computer Vision Center, UAB. He obtained the 2008 best Thesis award on Computer Science at Universitat Autònoma de Barcelona. He lead the Human Pose Recovery and Behavior Analysis Group at University of Barcelona. His research interests include, between others, machine learning, statistical pattern recognition, visual object recognition, and human computer interaction systems, with special interest in human pose recovery and behavior analysis.
\end{IEEEbiography}
\vspace{-2 cm}
\begin{IEEEbiography}[{\includegraphics[width=1in,height=1.25in,clip,keepaspectratio]{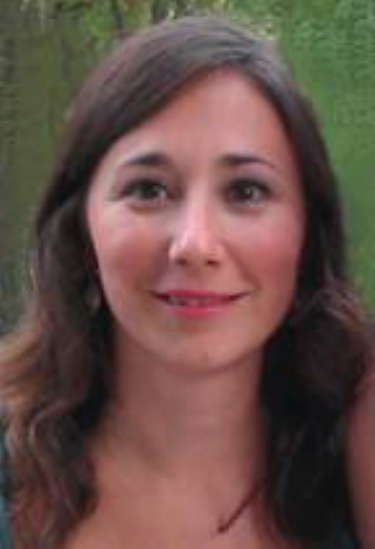}}]{Laura Igual}

Laura Igual received the degree in Mathematics from Universitat de Valencia in 2000. She developed her Ph.D Thesis in the program of Computer Science and Digital Communication at the Department of Technology of the Universitat Pompeu Fabra. She obtained her Ph.D Thesis in January 2006 and since then she is a research member at the Computer Vision Center (CVC) of Barcelona. Since 2009, she is a lecturer at the Department of Applied Mathematics and Analysis of the Universitat de Barcelona. She is a member of the Perceptual Computing Lab and a consolidated research group of Catalonia. Her research interests include medical imaging, with focus on neuroimaging, computer vision, machine learning, and mathematical models and variational methods for image processing.
\end{IEEEbiography}
\vspace{-2 cm}
\begin{IEEEbiography}[{\includegraphics[width=1in,height=1.25in,clip,keepaspectratio]{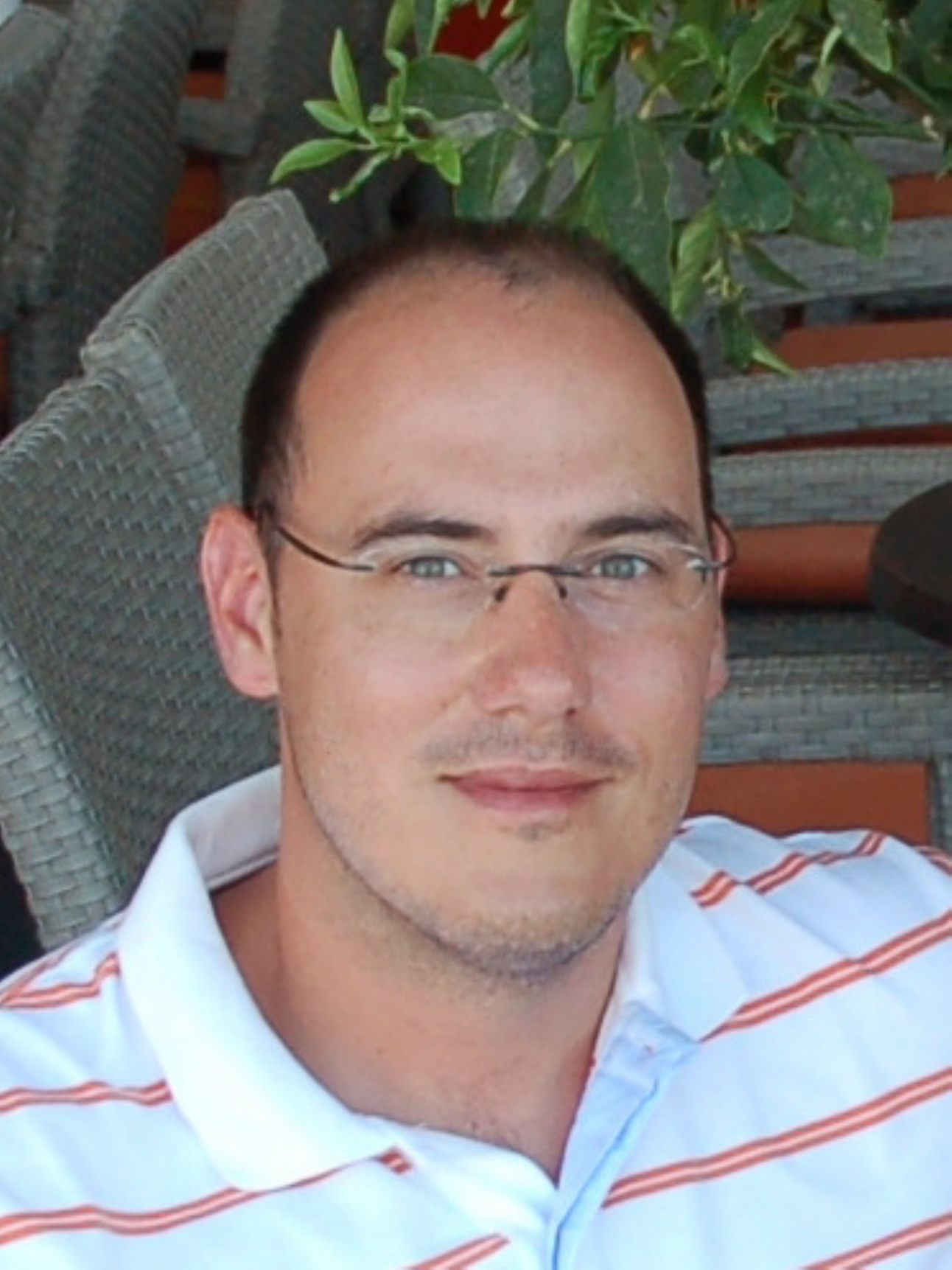}}]{Oriol Pujol}
Oriol  Pujol Vila obtained the degree in Telecomunications Engineering in 1998 from the Universitat Politècnica de Catalunya  (UPC). The same year, he joined the Computer Vision Center and the Computer Science Department at Universitat  Autònoma de Barcelona  (UAB). In 2004 he received the Ph.D. in Computer Science at the UAB on work in deformable models, fusion of supervised and unsupervised learning and intravascular ultrasound image analysis. In 2005 he joined the Dept. of Matemàtica Aplicada i Anàlisi at Universitat de Barcelona where he became associate professor. He is member of the BCN Perceptual Computing Lab. He has been since 2004 an active member in the organization of several activities related to image analysis, computer vision, machine learning and artificial intelligence
\end{IEEEbiography}
\vspace{-2 cm}
\begin{IEEEbiography}[{\includegraphics[width=1in,height=1.25in,clip,keepaspectratio]{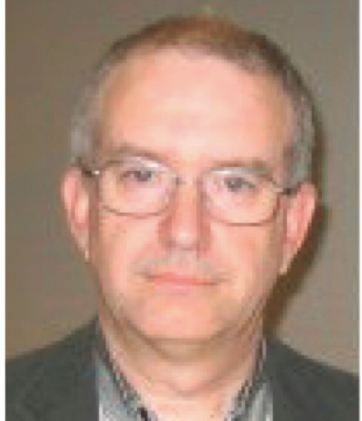}}]{Josep Moya}
Doctor in Medicine, Psychiatry and Psychoanalyst. He is with the Mental Health Department at Parc Taul\'{i} (Barcelona), he also is the leader of the Observatory of Communitarian Mental Health of Catalonia. He is a teacher in the Department of Social Wellness and Family of the Generalitat de Catalunya, and he also is teaching in the Center for Legal Studies and Specialized Training at the Department of Justice of the Generalitat de Catalunya. He is the president of CRAPPSI (Private Catalan Foundation for Research and Evaluation of Psychoanalytic Practice). Currently he leads a research project on the Impact of the Economic Crisis on the Mental Health of the Population. He has published several articles on ADHD.
\end{IEEEbiography}
\vspace{-2 cm}
\begin{IEEEbiography}[{\includegraphics[width=1in,height=1.25in,clip,keepaspectratio]{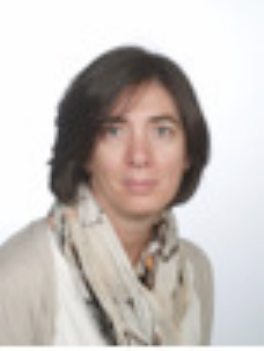}}]{Ver\'{o}nica Violant} obtained her Ph. D in Psychology from the Ramon Llull University. She is a tenured professor at University of Barcelona, currently at the Didactic and Educational Organization Department. She leads the graduate course on Pedagogics, Childhood and Disease at University of Barcelona. She is a member of the research group for Socio-educational Interventions in Childhood and Youth. Her research interests are hospital pedagogics. Concretely, in paediatrics and neonatology. She is author of various publications on the attentiveness on diseases in childhood and youth. In 2012 she was awarded with the Diamond Prize of research of the International Awards of the Eureka Sciences.
\end{IEEEbiography}
\begin{IEEEbiography}[{\includegraphics[width=1in,height=1.25in,clip,keepaspectratio]{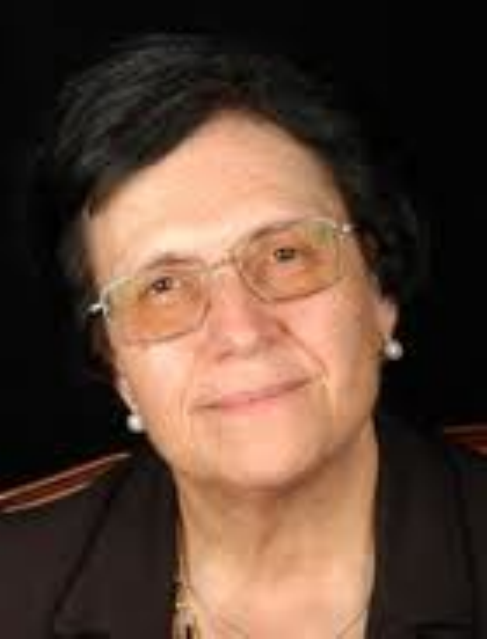}}]{Maria Teresa Anguera} obtained her  Ph.D in Philosphy and Humanities (Psychological section) at University of Barcelona. Maria Teresa Anguera also holds a Degree in Law from University of Barcelona. Maria Teresa is a distinguished professor at the Department of Behavioral Science Methodologies at University of Barcelona since 1986. Maria Teresa has a long teaching trajectory at University of Barcelona together with several research participations at foreign universities. Maria Teresa has advised several Ph.D dissertations and has publised more than 100 journal papers on psychology. She is an academic at the Spanish Royal Academy of Medicine. She has also been vice-rector of Scientific Politics at University of Barcelona. Since 2011 she is a member of the Steering Doctorate Committee.

\end{IEEEbiography}

\end{document}